\documentclass[10pt,journal,compsoc]{IEEEtran}

\ifCLASSOPTIONcompsoc
  \usepackage[nocompress]{cite}
\else
  \usepackage{cite}
\fi

\usepackage{array}
\usepackage{amsmath,amssymb,amsfonts}
\usepackage{algorithmic}
\usepackage{graphicx}
\usepackage{textcomp}
\usepackage{xcolor}
\usepackage{enumerate}
\usepackage{url}
\usepackage{multirow}
\usepackage{booktabs}

\begin{document}

\title{Modeling emotion in complex stories: the Stanford Emotional Narratives Dataset}


\author{Desmond~C.~Ong,~\IEEEmembership{Member,~IEEE~Computer~Society},
Zhengxuan~Wu, \protect\\
Tan~Zhi-Xuan, Marianne~Reddan, Isabella~Kahhale, Alison~Mattek, and Jamil~Zaki
\IEEEcompsocitemizethanks{
\IEEEcompsocthanksitem D. C. Ong is with the Department of Information Systems and Analytics, National University of Singapore, and with the A*STAR Artificial Intelligence Initiative,
Agency for Science, Technology and Research, Singapore \protect\\ 
E-mail: desmond.c.ong@gmail.com
\IEEEcompsocthanksitem Z. Wu is with the Department of Management Science and Engineering, Stanford University.
\IEEEcompsocthanksitem T. Zhi-Xuan is with the Department of Electrical Engineering and Computer Science at the Massachusetts Institute of Technology, and with the A*STAR Artificial Intelligence Initiative.
\IEEEcompsocthanksitem M. Reddan, I. Kahhale, and J. Zaki are with the Department of Psychology, Stanford University.
\IEEEcompsocthanksitem A. Mattek is with the Department of Psychology, University of Oregon.}
\thanks{Manuscript received February 15, 2019; revised August 11, 2019.}
}

%
%

\markboth{IEEE Transactions on Affective Computing, Manuscript ID}%
{Ong \MakeLowercase{\textit{et al.}}: Modeling emotion in complex stories: the Stanford Emotional Narratives Dataset}

\IEEEtitleabstractindextext{%
\begin{abstract}

Human emotions unfold over time, and more affective computing research has to prioritize capturing this crucial component of real-world affect. Modeling dynamic emotional stimuli requires solving the twin challenges of time-series modeling and of collecting high-quality time-series datasets. We begin by assessing the state-of-the-art in time-series emotion recognition, and we review contemporary time-series approaches in affective computing, including discriminative and generative models. We then introduce the first version of the Stanford Emotional Narratives Dataset (SENDv1): a set of rich, multimodal videos of self-paced, unscripted emotional narratives, annotated for emotional valence over time. The complex narratives and naturalistic expressions in this dataset provide a challenging test for contemporary time-series emotion recognition models. We demonstrate several baseline and state-of-the-art modeling approaches on the SEND, including a Long Short-Term Memory model and a multimodal Variational Recurrent Neural Network, which perform comparably to the human-benchmark. We end by discussing the implications for future research in time-series affective computing.

\end{abstract}

}

\maketitle
\IEEEdisplaynontitleabstractindextext
\IEEEpeerreviewmaketitle

\IEEEraisesectionheading{\section{Introduction}}

\label{sec:Introduction}

\IEEEPARstart{E}{motions} are an integral part of our everyday lives that dynamically color our experiences. For example, John may wake up feeling sad that he has to get out of his warm bed, then happy when he checks his phone and receives a nice email; and later frustrated when his bus to work arrives twenty minutes late. Our emotions vary dynamically over time, and are situated in the context of the day's events and our history of prior experiences \cite{ellsworth2003appraisal, ortony1988cognitive}.

As we open our homes, hospitals, and offices to artificial agents, our relationship with AI will become more personal. 
In order for these artificial agents to successfully co-exist with people, they will have to ``understand" our thoughts and emotions and react accordingly \cite{ong2015affective, ong2019computational}. 
The field of affective computing has made exciting progress in this direction, for instance, training artificial agents and algorithms to recognize emotions from faces \cite{sariyanidi2015automatic}, paralinguistics (e.g., pitch, prosody) \cite{schuller2013computational}, body gestures \cite{castellano2007recognising}, and language \cite{calvo2013emotions}. Newer approaches also involve integrating these types of cues into \emph{multimodal} judgments about the underlying affect \cite{zeng2009survey, poria2017review}.

A growing body of work in affective computing focuses on capturing and modelling the \textit{dynamics} of emotion as they unfold over time---what we refer to as \textbf{time-series emotion recognition}.
Specifically, we define time-series modeling as taking in temporally continuous input data and producing temporally continuous output, with an explicit consideration of how information is propagated over time. For instance, in order to engage in such inference, a social robot in conversation with its user would have to take in a continuous stream of sensor data, process them, and reason about their user's emotions \emph{over time}, perhaps after every second or after every sentence, as well as across many sentences in the conversation and across multiple conversations \cite{gunes2013categorical}.

Despite the progress that has been made in time-series emotion recognition in the past decade, the field is still far from affective robots that can understand human emotions in daily life. 
What is needed to achieve this ambitious goal? We suggest that the biggest barriers to overcome are due to (1) the inherent difficulty of building computational time-series models, and (2) the difficulty of collecting high-quality datasets. To address this first gap, we conduct a review covering different machine-learning-based approaches to time-series modeling (Section \ref{sec:Time-Series-Models}). We begin by discussing the most common time-series techniques in affective computing: deep neural network models, part of a broader class of \emph{discriminative} models. We also cover \emph{generative} time-series approaches, which are comparatively less popular within affective computing, but offer interesting modeling capabilities and hold exciting potential for emotion understanding.

We turn next to discuss the second gap: Researchers need high-quality time-series datasets on which to train models. These are expensive to construct, in terms of both the production of stimuli and the collection of time-series annotations of emotion and affective labeling \cite{schuller2014multimodal}. There are several existing time-series datasets that have been used by the affective computing community, mostly through the Audiovisual Emotion Challenges (AVEC), a series of challenges held annually since 2011 \cite{schuller2011avec}. AVEC is a large and collaborative multi-institutional effort that involves collating, curating, and releasing datasets, and has catalyzed much of the research in time-series affective computing. Every AVEC challenge to date involves producing time-series labels on a common dataset. 
The first two challenges \cite{schuller2011avec, schuller2012avec} had researchers predict valence over time on the SEMAINE dataset \cite{mckeown2012semaine}, which consists of recordings of volunteers interacting with a ``Sensitive Artificial Listener", an artificial agent programmed to respond in emotional stereotypes (e.g., happy and outgoing, or angry and confrontational \cite{schuller2011avec}).
The subsequent two AVEC challenges \cite{valstar2013avec, valstar2014avec} asked for predictions of valence and arousal on the AViD-Corpus, a series of recordings of volunteers performing several tasks like reading aloud storybook excerpts and describing the story behind a given picture (as in the Thematic Appreciation Test).
The fifth and sixth challenges \cite{ringeval2015av, valstar2016avec} involved predicting valence and arousal on the REmote COLlaborative and Affective interactions database (RECOLA; \cite{ringeval2013introducing}), which included pairs of individuals collaborating on a task via remote conferencing. 
Finally, the seventh and eighth challenges \cite{ringeval2017avec, ringeval2018avec} required predictions of valence, arousal and likability ratings on the Sentiment Analysis in the Wild (SEWA) dataset \cite{kossaifi2019sewa}, which also involved dyads discussing their views on a commercial that both individuals viewed. Unlike the previous three datasets, the SEWA dataset was collected ``in the wild" using participants' personal webcams rather than in a controlled lab environment. 
More recent challenges that involve predicting emotions or empathy over time include the 2018 OMG-Emotion \cite{barros2018omg} and the 2018 Affect-in-the-Wild challenge \cite{kollias2019deep}, both comprising collections of YouTube videos of spontaneous emotion displays, and the 2019 OMG-Empathy challenge, which had videos of a research volunteer listening to a confederate recount scripted emotional stories. Finally, several relevant time-series datasets were also published outside these challenges: The Belfast Induced Natural Emotion Database \cite{sneddon2011belfast} contains 1,400 clips of research volunteers performing tasks designed to elicit one of seven emotions (e.g., disgust: reaching into a box and touching cold spaghetti). The Affectiva-MIT Facial Expression Dataset (AM-FED; \cite{mcduff2013affectiva}) contains 242 videos of people at home watching an advertisement, and these videos were collected using their webcams. The Acted-Facial-Expressions-in-the-Wild--Valence-Arousal (AFEW-VA; \cite{kossaifi2017afew}) database contains 600 video excerpts from movies\footnote{Though we disagree that acted expressions are ``in the wild", as they do not occur naturalistically.}, annotated per-frame for valence and arousal.

The range of datasets we mentioned does not span the range of social interactions that arise in real life.  
In particular, previous datasets either tended to have a very constrained scope---such as interacting with the same agents or confederates (SEMAINE, OMG-Empathy), doing a fixed set of tasks (AViD, Belfast Induced Natural Emotion), or collaborating on a single task (RECOLA, SEWA)---or they tended to be too unconstrained---the OMG-Emotion, Affect-in-the-Wild, and AFEW-VA datasets contain emotion displays with no shared context. To fill this gap, we aimed to design a \emph{minimally-constrained context} that is both ecologically-valid and generalizable while still allowing for desired variability in emotional content and emotional expression. We settled on a context relevant to any conversational AI: first-person narrated personally-meaningful emotional stories\footnote{We note that one of the tasks in AViD also had participants tell personal stories, but the topics were assigned to the participant: ``best present" and ``sad event from childhood".}. In this manner, there is sufficient shared context in the dataset across participants, as each responded to the same prompt, as well as substantial inter-stimuli variability, especially in the content of the stories, on which we can train naturalistic emotion-recognition models. We call this new database of annotated videos of unscripted autobiographical emotional narratives the Stanford Emotional Narratives Dataset (SEND), and introduce the first release in Section \ref{sec:Dataset}. In Section \ref{sec:Modeling} we report the results of several baseline and state-of-the-art time-series modeling approaches on this dataset.

From this point onwards, we choose not to use ``continuous" to describe the time-series nature of the models or data. This is to avoid confusion with another potential meaning of ``continuous", which is to produce graded or dimensional outputs \cite{gunes2013categorical}. That is, instead of producing an emotion classification (e.g. \emph{happy} vs. \emph{sad} vs. \emph{neutral}) or a binary judgment (e.g. high or low valence), such models would predict a real-valued judgment on some interval or ordinal scale \cite{yannakakis2018ordinal}. We will stress here that the choice of a dimensionally-continuous output is an orthogonal modeling decision from dealing with temporally-continuous data, and hence we will not use ``continuous" to avoid ambiguity.

In the rest of this paper, we provide a review of time-series modeling, with a focus on affective computing (Section \ref{sec:Time-Series-Models}). We then introduce a novel naturalistic multimodal dataset consisting of unscripted emotional life stories (Section \ref{sec:Dataset}). In Section \ref{sec:Modeling}, we describe implementations of several baseline and state-of-the-art time-series approaches to modeling this dataset, and discuss the results in light of the modeling assumptions. Finally, we end with a discussion of how the field can extend these ideas to problems such as deploying these models in physical robots, and  building personalized and longitudinal affective computers that may interact with an individual over many sessions, potentially over a lifespan.

\section{Time-Series Models} \label{sec:Time-Series-Models}

In this section, we provide an overview of contemporary time-series approaches in affective computing. 
We do not cover linear models, such as autoregressive or moving average models traditionally used in econometrics and other fields: Rather, we focus on machine learning models that are more amenable to high-dimensional input data.

We use $X^k_t$ to denote the vector of input features for sequence $k$ at time $t$. This could be a vector of facial expression features or even multimodal features. We use $Y^k_t$ to denote the corresponding vector of outputs at time $t$, such as categorical labels of emotion classes or real-valued scores or probabilities. We use $X^k_{t_1:t_2}$ and $Y^k_{t_1:t_2}$ to denote a series of these inputs and outputs from times $t_1$ to $t_2$, inclusive. Given $n$ paired training sequences $\{\left(X^k_{1:T_k}, Y^k_{1:T_k} \right),\ 1 \leq k \leq n\}$ where $T_k$ is the final time point of sequence $k$, the goal is to train a model that can predict the sequence of outputs $Y^j_{1:T_j}$ given a new input sequence $X^j_{1:T_j}$, for some $j > n$. Without loss of generality, this new predicted sequence could also be an extension of a previously-observed sequence.

\begin{figure}[!bt]
\centering
\includegraphics[width=\columnwidth]{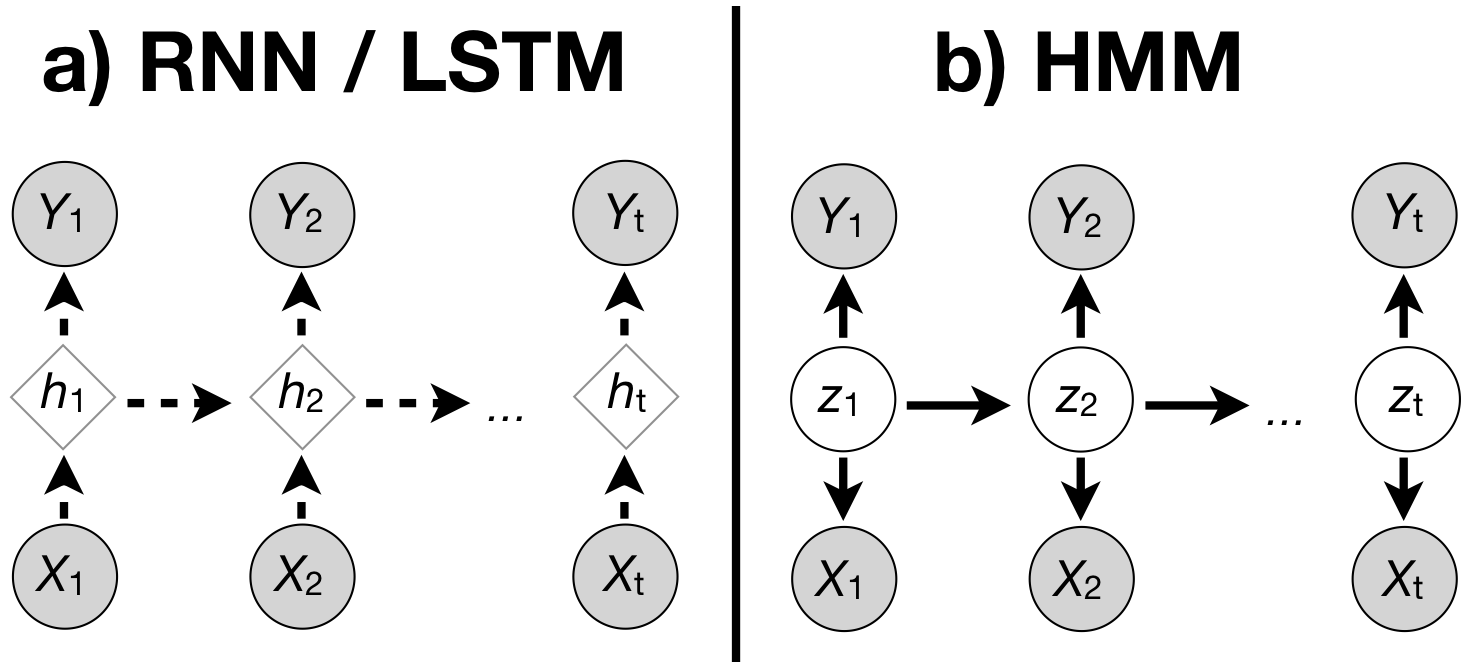}
\caption{Illustration of two common time-series models. We use conventional Bayesian Network notation, where circles represent random variables, shaded shapes represent observable quantities, and unshaded shapes represent latent quantities. We also use diamonds to represent deterministic values computed from random variables. (a) A popular discriminative time-series model, the Recurrent Neural Network (RNN). We use dashed lines to represent deterministic computations. Inputs $X_t$ (e.g., emotional expressions) are mapped onto hidden states $h_t$ to produce output labels $Y_t$ (emotion labels), and there is a recurrency between consecutive hidden states (Eqn \ref{eqn:RNN}). The discriminative approach finds the function that best discriminates the outputs given the inputs, modeling $P(Y|X)$. Long Short-Term Memory (LSTM) networks, are variants of RNNs where the hidden layers also includes ``memory" units that allow longer-range information dependencies. (b) A common generative time-series model, the Hidden Markov Model (HMM). Solid arrows represent causal influence. In the generative approach, there is some hidden (emotional) state $z_t$, which ``causes" people to display emotional expressions $X_t$ and also ``causes" observers to rate these as certain emotional states $Y_t$. The goal of the generative approach is to model the joint distribution $P(X,Y)$, in the case of the HMM, by invoking and marginalizing out latent variables $P(X,Y) = \sum_z P(X,Y|z)P(z)$.  
}
\label{fig:LitReview}
\end{figure}

%
%

\subsection{Discriminative Models}

Given a set of emotion outputs $Y_t$ and a set of input features $X_t$, one approach is to directly model how we can predict the output labels from the input features. Such \emph{discriminative} models \cite{ng2002discriminative} are widely used in machine learning for both classification (e.g., predicting an emotion category) and regression problems (e.g., predicting a real-valued number). Linear and Logistic regression, the Support Vector Machine/Support Vector Regression \cite{sun2016exploring}, Random Forest Classifiers \cite{kachele2015ensemble} and Deep Neural Networks like Convolutional Neural Networks \cite{fan2016video}, are amongst the most popular discriminative machine-learning models applied within (non-time-series) affective computing \cite{poria2017review, zeng2009survey}. 

A vanilla (standard) feed-forward neural network transforms inputs $X$ into outputs $Y$ via nonlinear transformations through intermediate, hidden layer(s) $h$. The most straightforward way to extend feed-forward neural networks to model time-series data is to allow the hidden layer at one time point to influence the hidden layer at subsequent time points. Adding such a ``recurrency" between hidden states results in an architecture known as the Recurrent Neural Network (RNN) \cite{williams1989learning}, shown in Fig. \ref{fig:LitReview}a. 
An RNN is a neural network in which the hidden state at time $h_t$ depends on the input features at that time $X_t$ and the hidden state at the previous time-point $h_{t-1}$, via some function $f$ with parameters $\theta$. The hidden states subsequently predict the outputs via $g$ with parameters $\phi$: 

\begin{align}
    h_t &= f_\theta \left(X_t, h_{t-1} \right) \nonumber \\
    Y_t &= g_\phi \left( h_t \right) \label{eqn:RNN}
\end{align}
In common parameterizations, $f_\theta$ and $g_\phi$ return a linear combination of their arguments filtered through a non-linear activation function (e.g., the hyperbolic tangent, the sigmoid, the softmax, or the Rectified Linear Unit (ReLU) functions). An example of a common formulation is:
\begin{align}
    h_t &= \tanh(W_X \cdot X_t + W_h \cdot h_{t-1}); \nonumber \\ 
    Y_t &= \text{softmax}(W_Y \cdot h_t) \label{eqn:RNN2}
\end{align} 
The weight matrices $W_X$, $W_h$, and $W_Y$ are shared across all time steps and learnt via stochastic gradient descent on the backpropagation of errors. 

One limitation of vanilla RNNs is that they do not readily capture long-range dependencies. Hochreiter and Schmidhuber \cite{hochreiter1997long} proposed adding memory units, or cells, within an RNN, which are able to ``remember" information over arbitrarily-long intervals. These Long Short-Term Memory (LSTM) networks have already become one of the most popular variants of the RNN, and we illustrate one variant in detail in Section \ref{sec:Modeling:LSTM}.

Many researchers have since used RNNs and their LSTM variants to recognize emotion from speech and from video. \cite{kahou2015recurrent, brady2016multi, khorrami2016deep} and \cite{kollias2019deep} all used a Convolutional Neural Network to learn hidden layer features from individual video frames, along with a recurrency between hidden layers at consecutive times---thus, combining the time-independent CNN with a RNN. Many others have used LSTMs to recognize emotions from video data. \cite{wollmer2008abandoning, eyben2010line} and \cite{wollmer2013lstm} were some of the earlier papers that worked on comparing multimodal LSTMs with Support Vector Regressions and other approaches for valence and arousal classification recognition on the SEMAINE dataset. This subsequently led to a surge of interest in applying LSTMs, especially to time-series emotion recognition on the AVEC 2015 \cite{chao2015long, chen2015multi}, AVEC 2017 \cite{huang2017continuous, chen2017multimodal}, AVEC 2018 \cite{zhao2018multi}, and OMG-Empathy 2019 \cite{tan2019Multimodal} challenges. Other noteworthy examples are \cite{pei2015multimodal}, who investigated bidirectional LSTMs (where there is another recurrence that goes backwards in time), \cite{wu2019attending} who combined neural attention mechanisms with LSTMs, and \cite{soleymani2014continuous} who built an LSTM with electroencephalography (EEG) input. These papers have collectively found that RNNs/LSTMs are a powerful model for time-series emotion recognition, whether they rely on extracted low-level features, or combined with features extracted using CNNs.

Discriminative approaches are, by and large, the most popular type of time-series approaches, because they provide a flexible approach that makes little assumptions about the nature of the data. At their heart, these approaches perform excellent pattern recognition, and find the best nonlinear functions that maps the input behavioral features to the output emotion via minimizing the error of the predictions of the model (also called the loss function). For tasks like emotion recognition from faces, deep approaches like Convolutional Neural Networks are by far the best performing state-of-the-art. One drawback, however of making less structural assumptions about the data is that these discriminative approaches, especially deep neural network approaches like LSTMs, tend to require larger amounts of data to learn and perform well.

There is another important modeling decision for such models: how to deal with asynchronous inputs. Multimodal time-series input often come in at different sampling frequencies, and discriminative approaches require some kind of binning to synchronize them \cite{gunes2005affect, snoek2005early}. One popular method (and the one that we use in this paper) is feature fusion, also called early fusion, where the input modalities are oversampled,  undersampled, or otherwise averaged, to a common sampling rate. 
This allows the multimodal features to be concatenated into a single feature vector within a given time window, to be fed into a model \cite{chao2015long, chen2015multi, brady2016multi}. A second way to achieve such ``synchronization" is decision fusion (or late fusion): This involves fitting a separate time-series model to each modality, operating at their own sampling frequencies. These individual models are then connected later in the computation to predict outputs \cite{huang2017continuous, dang2017investigating}.

%
%
\subsection{Generative Models}

A second class of time-series approaches instead focuses on modeling the causal structure behind the generation of the data \cite{ong2015affective, ong2019applying}. As we highlighted in the opening example, emotions dynamically vary over time, and cause behavior like emotional expressions. Thus, if we took a modeling approach that is more sensitive to the underlying emotional phenomena, we may be interested in explicitly writing out how, say, the emotions vary over time ($Y_t \! \rightarrow \! Y_{t+1}$), and how emotions cause emotional expressions ($Y_t \! \rightarrow \! X_{t}$). 
Generative models offer this flexibility along with their own share of modeling assumptions and challenges. More generally, generative models aim to model the joint distribution of the observed data, both the inputs $X$ and the outputs $Y$, or $P(X,Y)$. Indeed, the parameters in generative models are fit by maximizing the (log-)likelihood of the data under the model. By contrast, the discriminative models described in the previous subsection directly model the outputs given the features $P(Y|X)$, and are often trained by minimizing some loss function, which does not correspond directly to likelihood (see \cite{ng2002discriminative} for more discussion).

Let us illustrate this with a classic time-series generative model, the Hidden Markov Model (Fig. \ref{fig:LitReview}b). In this model, we posit that there is a latent (unobservable) variable $z_t$. This $z_t$ is a discrete, categorical variable 
(e.g., a discrete emotion category like \emph{happy} or \emph{sad}): it could also be some unknown ``state of the world" that the modeller may be agnostic about labeling. First, the latent variable at the current time step $z_t$ ``causes" both the input features $X_t$ and the output labels $Y_t$ via an emission function or emission model $z_t \! \rightarrow \! (Y_t; X_t)$. The model's emission probabilities encode how observations are ``emitted" from the hidden states. 
Second, the latent variable at the current time step $z_t$ changes at the next time step $z_{t+1}$ via a transition function $z_t \! \rightarrow \! z_{t+1}$ with transition probabilities governing how one hidden state may transition to another. 
The $X_t$'s and the $Y_t$'s are only connected via the $z_t$'s, and each $z_t$ is only influenced by the $z$ at the preceding time-point.

The HMM allows one to set priors on both the transition and emission models. For example, one might have a theory that emotions tend to be ``sticky" over the time-scale of the time steps \cite{kuppens2010feelings}, so the emotional state $z_t$ would likely be similar to the preceding state $z_{t-1}$. Alternatively, emotion A may be more likely to precede emotion B than emotion C \cite{sudhof2014sentiment}: These could all be set in the transition model via weights in a multinomial distribution. These priors are updated after observing the data. More generally, we can define parameterized distributions, and find the parameters $\theta$ that maximize the probability of the data under the model:
\begin{align}
    z_t &\sim P_\theta(z_t | z_{t-1}) \nonumber \\
    X_t &\sim P_\theta(X_t|z_t) \nonumber \\
    Y_t &\sim P_\theta(Y_t|z_t) \nonumber \\
    \theta^* &= \arg\max_{\theta} P_\theta(X_1,\ldots,X_T, Y_1,\ldots,Y_T)
\end{align}

Hidden Markov Models have been used for many years to recognize time-series emotions, especially from speech. \cite{schuller2003hidden, nogueiras2001speech} and \cite{jiang2004speech} all explored using HMMs to classify speech into discrete emotion categories. \cite{wagner2007systematic} did a more systematic investigation of how various parameters of HMMs (e.g. number of states or mixtures per state, input lengths) impact their performance at recognizing emotions in speech. The latent variable in a HMM can also capture different types of variability: for example, emotion dynamics within an utterance, versus emotion dynamics within a conversation across multiple utterances. \cite{metallinou2012hierarchical} modelled exactly these two levels of emotion dynamics using a HMM with two hierarchical layers of latent variables. \cite{cohen2000emotion} also applied a multilevel HMM to recognize emotions from sequences of facial expressions. We implement a HMM as a baseline model in Section \ref{sec:Modeling:HMM}

Researchers have also tried other similar generative models to emotion recognition. For example, a Kalman Filter is similar to a HMM with one main difference being that the hidden states are real-valued instead of categorical: 
\cite{somandepalli2016online} applied Multimodal Kalman Filters to recognize valence and arousal over time on the AVEC 2016 challenge. Working on the same dataset, \cite{atcheson2017gaussian} applied a Gaussian Process Regression model, which is similar to a Bayesian Regression in that they assume a generative (Gaussian) process over the parameters of a regression model. 
\cite{dang2017investigating} also used a Gaussian Process Regression, as well as a Gaussian Mixture Regression (which assumes that the model parameters are a result of a ``mixture" or combination of multiple Gaussians) to recognize valence and arousal from multimodal cues on the AVEC2017 dataset.

Compared to discriminative approaches, generative approaches make more assumptions about the underlying structure of the data, such as which variables ``cause" which other variables and how. These modeling assumptions provide an inductive bias \cite{schulz2017compositional, lake2017building} that helps models to learn faster with less data. Generative models also allow the model to learn different sources of variability. For example, by using hierarchical latent-variable models \cite{metallinou2012hierarchical}, we could potentially learn general emotion-cue mappings (e.g., people tend to smile like so when happy) as well as person-specific mappings (Bob tends to smile like \emph{that} when happy). 

One drawback, however, is that generative models tend to make strong simplifying assumptions: HMMs for example, are defined on discrete states with simple transition functions, while Kalman filters similarly assume linear (and Gaussian) dynamics. This limits their ability to express complicated models, compared to discriminative approaches that can theoretically learn very complex functions.
Generative models also tend to be more computationally expensive to train. Inference in these models is often NP-hard and tractable only in simple models, and so many models rely on various approximate-inference algorithms. Thus, generative models face a dilemma: They tend to either be (i) too simple to sufficiently capture real-world variability, or (ii) too complex for fast, efficient inference.

\subsubsection{Integrating discriminative and generative approaches}

Fortunately, this is becoming less of a problem. 
In recent years, researchers have developed models that merge the benefits of the discriminative and generative approaches, for example, by using techniques from deep learning to produce more efficient approximate-inference algorithms. In non-time-series domains, the Variational Autoencoder \cite{kingma2013auto} has become a popular and flexible deep generative model---a generative model parameterized by neural networks, and where inference in the model can be approximated by maximizing a variational lower bound on the log-likelihood of the model. \textit{Variational inference} \cite{hoffman2013stochastic, blei2017variational} thus approximates the computationally-expensive inference problem by replacing it with a less-computationally-expensive optimization problem. Indeed, by parameterizing generative models with neural networks, one could learn arbitrarily complex functions linking the latent variables with the data. We recently proposed \cite{ong2019applying} that such deep generative approaches allow affective computing researchers to leverage the advantages of both generative and discriminative approaches, and illustrate with several (non-time-series) examples using VAEs and their variants. 

Within time-series modeling, there are a handful of promising examples of such integration of deep and generative models. For example, using a Deep Markov Model \cite{krishnan2017structured, archer2016black, zhixuan2020factorized}, one can parameterize the generative edges in the generative model (e.g., the emission and transition functions) using neural networks as in Eqn. \ref{eqn:RNN2}. In another example, \cite{chung2015recurrent} and \cite{bayer2014learning} both introduced a latent variable into an RNN to help it model different sources of variability (e.g. inter-subject variability) in the data. Within affective computing, \cite{wollmer2010combining} combined an LSTM and a Dynamic Bayesian Network to extract word-level linguistic features for predicting valence and arousal.

We hope that in due course, these contemporary hybrid techniques will improve by leveraging strengths of both approaches, and subsequently be adopted within the affective computing community. In Section \ref{sec:Modeling:VRNN}, we present an implementation of a modified Variational Recurrent Neural Network \cite{chung2015recurrent} that combines deep and generative approaches.

Finally, we end our review by mentioning one more alternative class of models: event-based models such as point-process models \cite{xiao2017modeling, du2016recurrent, linderman2014discovering, qin2017event}, which aim to model the time and intensity of \textit{events} and their impact on a dependent variable. Although event-based approaches are not common within affective computing, one notable and recent example is \cite{wataraka2018speech}, who proposed an event-filter model to predict valence and arousal over time from speech events. In their model, a vocal event $j$, occuring at times predicted by $\varphi_j(t)$, produces an emotional ``response" $h_j(t)$. For example, if event $j$ denotes laughter, then $\varphi_j(t)$ captures all the occurrences of laughter in the signal, and $h_j(t)$ represents the change in emotional valence signalled by a single laughter episode (perhaps, a sharp increase, followed by some decay back to baseline). Then, the emotional signal $Y(t)$ is then proportional to the sum of the convolution $h_j(t) \bigoplus \varphi_j(t)$ across all events $j$. They tested their event-filter model on the AVEC 2018 dataset, and it performed better than the audio-channel-only baselines for the AVEC 2017 and 2018 challenges \cite{wataraka2018speech}. While we do not go further into event-based models in this paper, we do think it offers an alternative approach to modeling time-series emotions, which should be explored more in future research. For example, by contrast to the approach of recognizing emotions from emotional expressions (which \cite{wataraka2018speech} still employ), these event-based models could allow modeling emotions as arising from the subjective appraisals of discrete events, as in Appraisal Theory \cite{ellsworth2003appraisal, ortony1988cognitive}.

\section{The Stanford Emotional Narratives Dataset (SEND)} \label{sec:Dataset}

In order to build affective computers that can understand human emotions in real life, we need high-quality time-series datasets with naturalistic emotion expressions. Here, we introduce the first  version of the Stanford Emotional Narratives Dataset (SENDv1). The SENDv1 consists of video clips of people recounting important and emotional life stories. These unscripted narratives capture spontaneous naturalistic emotional expressions as well as complex semantic content. These stories also show diverse emotional ``trajectories", and thus provide a rich dataset for time-series modeling. Information regarding the SENDv1 is available at \url{https://github.com/StanfordSocialNeuroscienceLab/SEND}.

We refined the experimental protocol \cite{ong2017dissertation} for the collection of the SENDv1 following our previous work \cite{zaki2008takes, devlin2016tracking}. All experiments were approved by the Stanford University Institutional Review Board. 
Participants (``targets") were recruited from a suburban community on the West Coast of the United States. They were brought into the lab and told to think about the three most positive and three most negative events that they would feel comfortable sharing in front of a video camera. Recording was self-paced: The experimenter left the target alone in the room, and targets were allowed to talk for as long as they wanted about each event. After targets finished recording the videos, they were asked to fill out several trait, personality, and demographic surveys. During this time, the experimenter processed the videos by transferring them from the camcorder onto the computer and prepared the next part of the experiment.

After targets finished the surveys, they were then showed each video that they recorded. While watching each video, they were asked to rated how they felt as they were telling their story. These valence ratings were collected using a visual analog scale divided into a hundred points, ranging from ``Very Negative" (-1) to ``Very Positive" (+1). The ratings on the scale were sampled every 0.5s. 
Many previous studies have used similar continuous rating dials, scales, or joysticks \cite{levenson1983marital, ruef2007continuous, douglas2007humaine, cowie2012tracing, kollias2019deep} to provide continuous valence ratings of videos.
Finally, after watching all their videos and making ratings, targets were asked to give consent for us to use the videos in future experiments. The subset of video clips selected for the SENDv1 were all consented for research use.

Video and audio were captured using a consumer-grade camcorder (Canon VIXIA HF R62) recording in high-definition at 30 frames per second, although videos were later downsized to 480x270 before collecting observer ratings and analysis. 
We specifically designed the video collection to be as ``clean" as possible, both for people who watch the videos and for machine learning models: Targets did not have to wear any headphones, and only a minority of videos had visible physiological sensors\footnote{We also collected physiological measurements (heart-rate and galvanic skin response) using a Biopac MP150, although we do not analyze these in this paper. In some videos, the heart-rate sensors placed just under the collarbone were visible.}.
Targets were seated in front of a black backdrop to standardize the background and remove any distracting objects. We positioned the camera to capture targets' faces head-on, and down to their shoulders (See Fig. \ref{fig:ratingParadigm}).

\begin{figure}[!bt]
\centering
\includegraphics[width=\columnwidth]{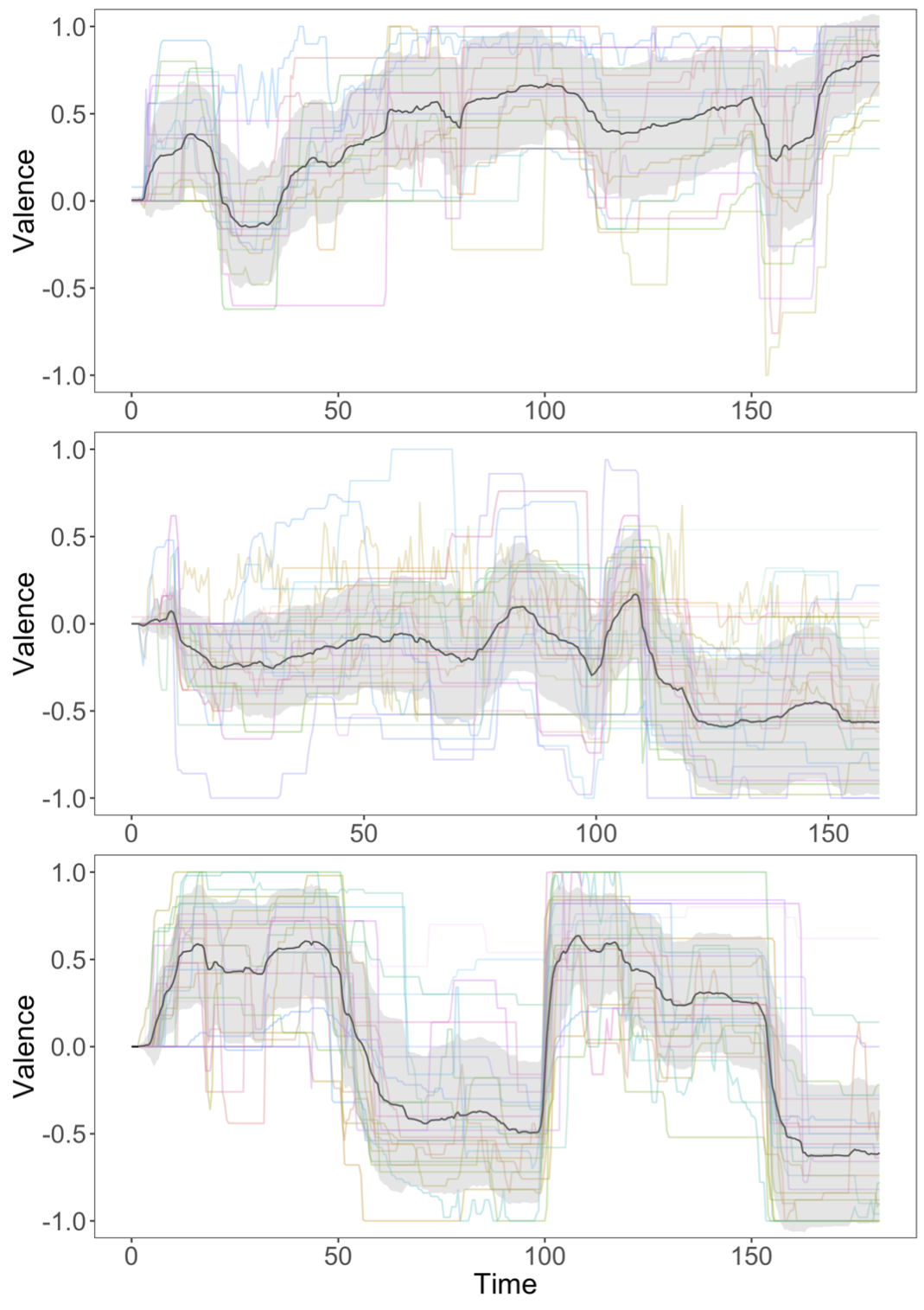}
\caption{Three example videos from the SENDv1. We collected independent observer ratings of the target's valence over time, which ranged from Very Negative (-1) to Very Positive (+1). Each colored line represents an individual observer's rating, and the black line represents the Evaluator Weighted Estimator of the observer ratings, along with its standard deviation. 
Top: A positive video, from our Test set, describing the buildup to receiving a puppy as a present. Middle: A negative video, from our Validation set, describing getting injured during a tournament season. Bottom: A mixed video, with both positive and negative segments, from our Training set, describing a drawn-out romantic breakup.
}
\label{fig:video_examples}
\end{figure}

We selected a subset of 193 clips containing 49 unique targets\footnote{Targets were 33\% White or Caucasian, 12\% East Asian, 8\% Hispanic or Latino/a, 6\% South Asian, 4\% Black or African American, and 2\% Middle Eastern. 35\% reported being of mixed heritage or ``Other".}. This set was chosen such that: (i) the target's face was always in the camera, (ii) the clips did not contain sensitive content (e.g. mental health, suicide), and (iii) the clips were emotional, and had some narrative flow (rather than stream of consciousness or rambling). The clips were also cropped for length, such that the final clips lasted on average 2 minutes 15 seconds (for a total of 7 hrs 15 mins). Targets in this subset talked about positive events like receiving a puppy as a surprise present (Fig. \ref{fig:video_examples}, top), successfully putting on a theatrical production, or going on a memorable vacation; to negative events like getting injured during a tournament season (Fig. \ref{fig:video_examples}, middle), witnessing parents fighting, or having a loved one pass away. Targets also described events that had both strongly positive and negative components, which we loosely term ``mixed" events, such as a long drawn-out romantic breakup with both ups and downs (Fig. \ref{fig:video_examples}, bottom). 
These unscripted narratives capture natural variation in emotion expression as the target is speaking, which is crucial in training affective computing models.
This dataset contains a rich sampling of emotional life events that people encounter in day-to-day life. We should also mention that, given its nature, we can keep building upon this dataset; We plan to supplement future versions of the SEND to encompass a wider variety of events, as well as targets from different racial, cultural, demographic, and socio-economic backgrounds.

\subsection{Independent Observer Ratings}
\label{sec:observerLabels}

\begin{figure}[!bt]
\centering
\includegraphics[width=.9\columnwidth]{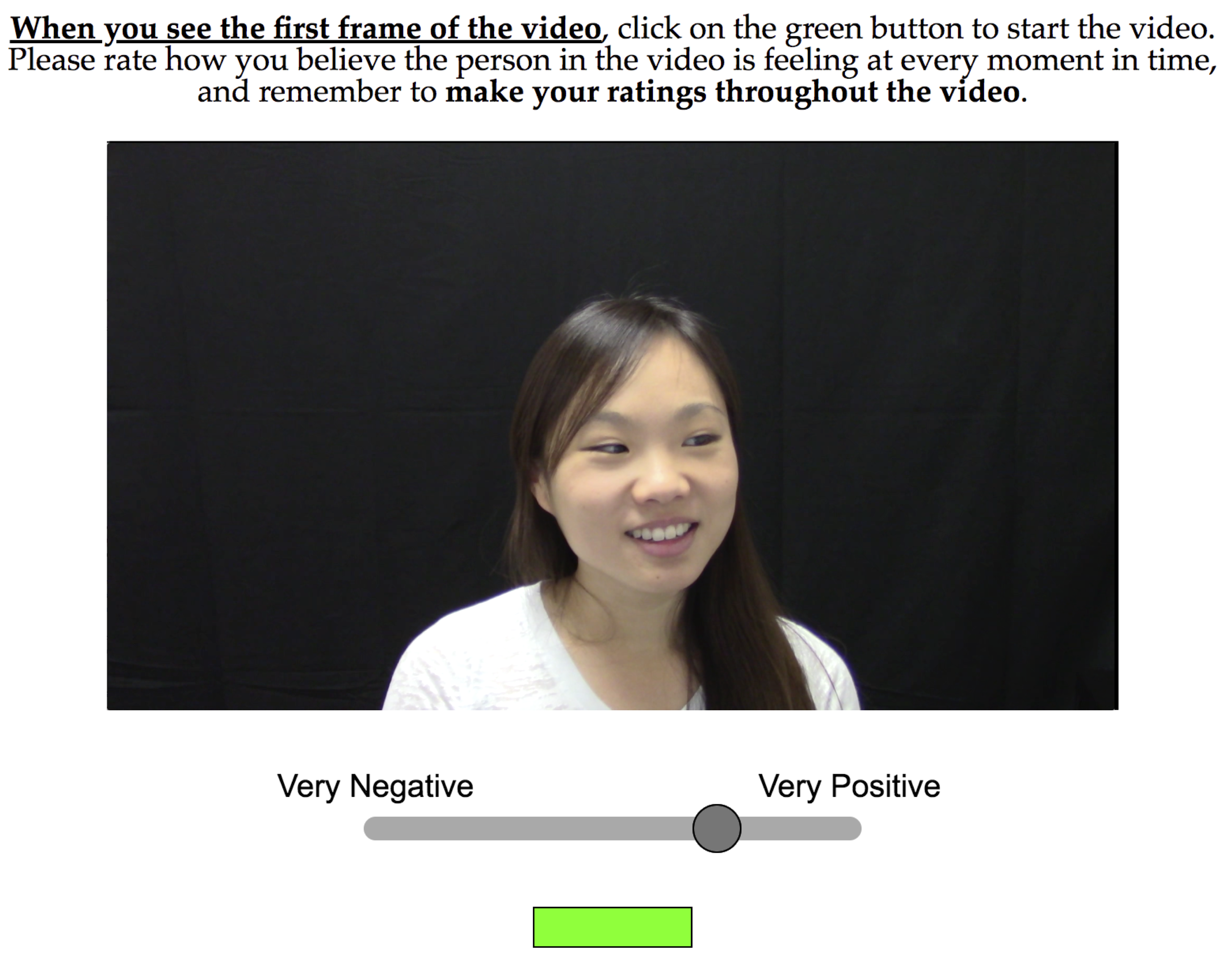}
\caption{Paradigm used to collect observer ratings. Observers used a visual analog scale from ``Very Negative" to ``Very Positive", and dragged the slider as the video was playing, to rate the target's valence. Videos captured targets' faces and shoulders against a clean, black backdrop.
}
\label{fig:ratingParadigm}
\end{figure}

As described above, we collected targets' self-reported valence ratings as labels for the videos. For the purposes of building an affective computer to recognize emotions, however, targets have access to more information than a computer could have---for example, targets know how the story is going to end even before the story begins, and targets have access to internal (visceral) cues of emotion. By contrast, an affective computer only has access to observable, externally-perceivable cues. 
Because of this consideration, we decided to additionally collect a large amount of independent ratings by recruiting a separate group of participants (``observers"). 
These independent ratings offer a different type of rating: that of \emph{externally-perceived} emotions. Arguably, externally-perceived emotions---done with only externally-observable cues and without hidden information such as memory or subjective feelings---is the goal of an affective computer. In this section, we describe the collection of the independent observer ratings, and which we use in the remainder of the paper. We will release both the targets' self-reported ratings and the observer ratings with the SEND, and researchers can choose which to use depending on their scientific hypotheses, but we do not discuss the target ratings in the remainder of the paper.

We recruited participants (``observers") on Amazon Mechanical Turk to watch these videos clips and provide ratings of how the target in the video felt along the valence dimension. Observers saw each video along with a continuous sliding scale underneath (Fig. \ref{fig:ratingParadigm}), and were asked to rate, using their mouse, how they thought the target was feeling as they were speaking in the video (and not how the target may have been feeling during the event they were describing). Observers were reminded to move the scale as the target is speaking to continually reflect the target's emotions. The analog scale was divided into 100 points and sampled every 0.5s.

Due to the complex nature of the stimuli, we aimed to get a large number of ratings ($>$20) per video for greater reliability. Hence, we recruited 700 observers, who each watched 8 videos. To ensure that observers were paying attention, we included two comprehension checks per video, which were True/False questions pertaining to the content of the video. Overall, observers got both attention check questions correct on 82\% of trials, one question correct on 15\% of trials, and zero of two correct only on 2\% of trials. We excluded trials on which observers answered zero or one questions correct, as well as on trials on which observers made no rating changes, resulting in a total of 3955 rating vectors, or an average of 20.5 rating vectors per video.

To calculate the ``gold-standard" valence labels, we used the Evaluator Weighted Estimator (EWE \cite{grimm2007primitives}), which provides an elegant formulation for weighting each observer's ratings by how well they correlate with the (unweighted) average of the ratings. 
Specifically, if observer $j$ provides rating $r^j_{1:T_k}$ for video $k$ of length $T_k$, and if the corresponding averaged rating of all raters is $\overline{r_{1:T_k}}$, then we can define a weight for observer $j$ (on video $k$), $w^j$, as the correlation between $r^j_{1:T_k}$ and $\overline{r_{1:T_k}}$. The EWE for video $k$ is then given by a weighted sum of the individual ratings:
\begin{align}
    w^j &= \text{Correlation}(r^j_{1:T_k}, \overline{r_{1:T_k}}) \\
    r^{\text{EWE}}_t &= \frac{1}{\sum_j w^j} \sum_j w^j r^j_t \quad (\text{for } 1\leq t \leq T_k) \label{eqn:EWE}
\end{align}

\subsection{Dataset Partitions}
We divided the videos into three partitions: \textbf{Training} (60\% of the dataset, 114 videos from 29 targets, 4 hrs 20 mins long), \textbf{Validation} (20\%, 40 videos from 10 targets, 1 hr 29 mins long) and \textbf{Test} (20\%, 39 videos from 10 targets, 1 hr 26 mins long) sets. See Table \ref{tab:datasetPartitions}. These partitions were done by target, so a particular target would only appear in one of the three partitions; This forces our models to learn to generalize to novel targets. We designed the partitions to have the same: (i) ratio of female vs. male gender presentation ($\chi^2(4)=.02, p=.99$, no gender non-conforming or ambiguous individuals were part of the dataset), (ii) mean video duration ($F(2,190)=.20, p=.82$), and (iii) ratio of positive/negative/mixed videos.  

\begin{figure}[!bt]
\centering
\includegraphics[width=\columnwidth]{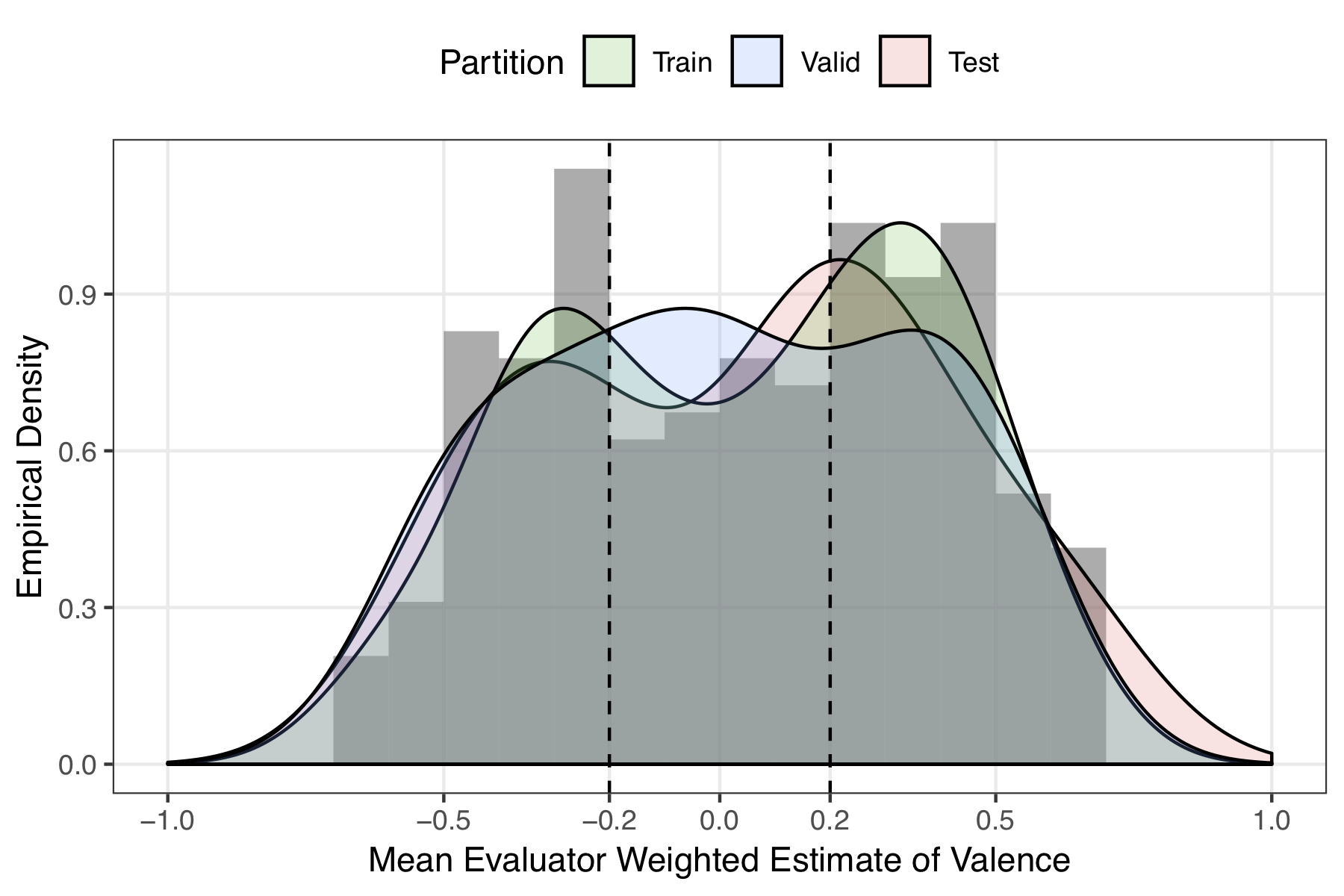}
\caption{Histogram of the mean EWE ratings (Eqn. \ref{eqn:EWE}) for each video. The grey histogram at the back reflects the distribution across the entire SENDv1, while the overlaid density distributions show the statistically-similar distribution of valences across the three partitions. The vertical dashed lines indicate our cutoffs for defining ``positive", ``mixed", and ``negative" videos.}
\label{fig:histogram}
\end{figure}

For the purposes of balancing the distribution of valences among the partitions, we defined ``positive" videos as those having a mean EWE rating of more than 0.2 (on a -1, Very Negative to 1, Very Positive scale). We similarly defined ``negative" videos as those having a mean EWE rating of less than -0.2, and ``mixed" videos as falling in between. As the videos were chosen to have meaningful emotional content, having a mean EWE around 0 suggests that there were both positive and negative segments in the video (See Fig. \ref{fig:video_examples}, bottom), rather than no emotional content. These cutoff values (of -0.2, 0.2) were chosen after looking at the distribution of mean EWE ratings, in Fig. \ref{fig:histogram}. The three partitions have a statistically similar ratio of positive to negative to mixed videos ($\chi^2(4)=.16, p=.99$). Overall across the whole dataset, there tends to be more positively-valenced ($39\%$) videos than mixed ($28\%$) and negative ($33\%$) videos, although this is not statistically different from a uniform split ($\chi^2(2)=3.8, p=.15$).

\begin{table}[ht]
    \centering
    \begin{tabular}{c|ccc|c}
         & Training & Validation & Test & Total \\
        \midrule
        \# Targets & 29 & 10 & 10 & 49 \\
        \# Female (\%) & 18 (62\%) & 6 (60\%) & 6 (60\%) & 30 (61\%) \\
        Mean Age (SD) & 24.8 (9.6) & 23.2 (4.6) & 21.1 (3.0) & 23.7 (7.9) \\
        \midrule
        \# Videos & 114 & 40 & 39 & 193 \\
        Total Length/s & 15622 & 5337 & 5186 & 26145 \\
        Avg Length/s (SD) & 137 (42) & 133 (37) & 133 (45) & 135 (41) \\
        \midrule        
        \# Pos. Vids (\%) & 46 (40\%) & 15 (38\%) & 15 (38\%) & 76 (39\%) \\
        \# Neg. Vids (\%) & 37 (32\%) & 13 (33\%) & 13 (33\%) & 63 (33\%) \\
        \# Mix. Vids (\%) & 31 (27\%) & 12 (30\%) & 11 (28\%) & 54 (28\%) 
    \end{tabular}
    \caption{Summary statistics of the 60:20:20 Training/Validation/Test partitions.}
    \label{tab:datasetPartitions}
\end{table}

\subsection{Model Evaluation}

We use the Concordance Correlation Coefficient (CCC \cite{lin1989concordance}) as the metric to compare our models' predictions for a time-series video with the gold-standard ratings. The CCC has been used in previous affective computing studies and challenges \cite{valstar2016avec, ringeval2017avec}. Intuitively, the CCC captures the expected discrepancy between the two vectors, compared to the expected discrepancy if the two vectors were uncorrelated. The CCC for two time-series vectors $X$ and $Y$ is:
\begin{align}
    \text{CCC}\left(X,Y\right) 
    &\equiv \frac{2 \text{ Corr}(X,Y) \sigma_X \sigma_Y}{\sigma_X^2 + \sigma_Y^2 + \left( \mu_X - \mu_Y \right)^2}
\end{align}
where $\text{Corr}(X,Y) \equiv \text{cov}(X,Y)/(\sigma_X \sigma_Y)$ is the Pearson correlation, and $\mu$ and $\sigma$ denotes the mean and standard deviation respectively. Like the Pearson correlation, the CCC measures agreement: $+1$ means that the two time-series are in perfect agreement and $0$ means that they are uncorrelated. The CCC also penalizes bias in the model's predictions via the $\left( \mu_X - \mu_Y \right)^2$ term in the denominator.

\section{Modeling} \label{sec:Modeling}

In this section, we present several time-series approaches to model valence ratings on the SENDv1. We implement:
\begin{itemize}
    \item a baseline (non-time-series) discriminative model, a Support Vector Regression (SVR)
    \item a baseline generative model, a Hidden Markov Model (HMM)
    \item a state-of-the-art discriminative Long Short-Term Memory (LSTM) model
    \item and a state-of-the-art (deep) generative Variational Recurrent Neural Network (VRNN) model.
\end{itemize} 
As is conventional practice, we train our models only on the Training Set, and use the models' performance on the Validation set to choose model hyperparameters (e.g., learning rate). We then use these optimized settings to report results on the Test set.
In addition to reporting mean results, we also report \emph{standard deviations} (SD): This is to show the variability in model performance across the different videos in a particular partition of the dataset (e.g. mean $\pm$ SD across all videos in the Test set). 
Although reporting SDs or other statistics is not yet commonplace in Machine Learning, we note that this is starting to change in recent years. 
The code for our models, written in Python, can be found at: 
{ \footnotesize \url{https://github.com/desmond-ong/TAC-EA-model}}.

\subsection{Human Benchmark}
\label{sec:Modeling:HumanBenchmark}

First, we wanted to establish how human observers perform on this task. This serves two purposes: First, it gives readers an intuition as to how difficult this task is. Second, it provides a quantitative benchmark with which to compare our modeling results in the next few sections.

We sought to calculate how well each individual observer $j$'s rating tracks the ``gold-standard" EWE (Eqn. \ref{eqn:EWE}), but because the EWE rating contains observer $j$'s rating, we calculated the CCC of $j$'s rating with an EWE that \textbf{has $j$'s rating subtracted out}. If $\mathcal{J}_k$ denotes the set of observers for video $k$ (of length $T_k$), $r^j_{1:T_k}$ denotes observer $j$'s ratings and $r^{\text{EWE}}_{1:T_k}|^{\mathcal{J}_k \setminus j}$ the EWE of all the other observers (minus $j$), then the mean human CCC on video $k$ is:
\begin{align}
    \overline{\text{CCC}}_k = \frac{1}{|\mathcal{J}_k|} \sum_{j \in \mathcal{J}_k} \text{CCC}\left( r^j_{1:T_k}, 
    r^{\text{EWE}}_{1:T_k}|^{\mathcal{J}_k \setminus j} \right) \label{eqn:humanCCC}
\end{align}
where $|\mathcal{J}_k|$ is the number of observers for video $k$.

Using Eqn. \ref{eqn:humanCCC}, the mean and standard deviation of observer CCC on the \textbf{Training set} was $.53 \pm .13$, the mean (and SD) observer CCC on the \textbf{Validation set} was $.47 \pm .15$, and finally, the mean (and SD) observer CCC on the \textbf{Test set} was $.50 \pm .12$.\footnote{The human-benchmark Train Set CCCs are significantly higher than those on the Validation Set ($p$=$.03$), but the human CCCs on the Test Set are not significantly different from either the Train or the Validation ($p$'s$>.27$). We do not think this is a problem with balancing; if anything, it means our experiments are more conservative as the Validation videos may be more challenging, even for humans.}

\subsection{Feature Extraction}

To facilitate comparison across the different model types, we chose to extract features from all the modalities and combine them into a multimodal input feature vector---This is also known as feature fusion or early fusion.

\textbf{Audio Features}. We used openSMILE v2.3.0 \cite{eyben2013opensmile} to extract the extended GeMAPS (eGeMAPS) set of 88 parameters recommended by \cite{eyben2016geneva}. Features were extracted for every 0.5-second window.

\textbf{Text Features}. We commissioned professional transcripts from a third-party company: These transcriptions were done manually with the aid of specialized annotation software to start, pause, and rewind the videos, but no automatic speech recognition software was used by the company. After receiving the text transcripts, we then used forced alignment\footnote{https://github.com/ucbvislab/p2fa-vislab} to assign timestamps to individual words. We used 300-dimensional GloVe word embeddings \cite{pennington2014glove} as a representation for each word. Features for each 5-second time window were then computed by averaging the word embeddings that occurred within each window.

\textbf{Visual Features}. We used the Emotient software by iMotions\footnote{https://imotions.com/emotient/} to extract 20 Action Units \cite{ekman1978manual} for each frame (30 per second).

To synchronize all three modalities with the ratings, all features were resampled to a common time window of 0.5 seconds before being fed into our models.

\subsection{Baselines}

\subsubsection{Support Vector Regression}
\label{sec:Modeling:SVR}

Following recent dataset papers that perform time-series valence prediction \cite{kossaifi2017afew, kossaifi2019sewa, mollahosseini2017affectnet}, we used Support Vector Regression (SVR) with a linear kernel as a baseline. SVR adapts the widely used Support Vector Machine (SVM) for use in regression tasks \cite{drucker1997support}, finding the hyperplane that best explains the data while allowing for a certain margin of error. Ratings were predicted from the inputs for each time window separately, and then smoothed using a simple moving average across 5 time windows (2.5s). We used the \texttt{scikit-learn} implementation of SVR \cite{pedregosa2011scikit}, and we cross-validated over multiple margins of error (0.05, 0.1, 0.15, 0.2) and error penalty terms ($10^{-3}$, $3 \times 10^{-3}$, ..., $3 \times 10^{2}$, $10^3$).

As we might expect, the baseline SVR does not do so well on this task, with the maximum performance it achieves is a CCC of .07 $\pm$ .13 on the Validation Set and  .08 $\pm$ .16 on the Test Set (see Table \ref{tab:SummaryOfResults} for a summary of all the model results). 
This poor performance is likely due to two reasons: First, SVR is not designed to handle time-series. We treated each time-step as an independent example, which is a poor assumption in such correlated video data. Second, given the complexity of our input features, using SVR with a linear kernel is unlikely to capture the relevant similarities between different input examples. This amounts to using a linear model on a non-linear regression problem, leading to poor prediction results.

\subsubsection{Hidden Markov Models}
\label{sec:Modeling:HMM}

Hidden Markov Models (HMMs) have been widely used for emotion recognition from speech \cite{schuller2003hidden}, facial \cite{cohen2003facial,gunes2008automatic}, and audio-visual data \cite{nicolaou2010audio,ozkan2012step}. In standard HMMs, the hidden states have to be discrete, so we adopted the approach in \cite{ozkan2012step} and discretized the valence ratings into multiple bins of equal sizes, treating each valence bin as the hidden emotional state to be recognized. We used multivariate Gaussian mixture models for the emission distributions of our HMM, with diagonal covariances for each Gaussian component. We fit the HMM via supervised learning using the \texttt{pomegranate} library \cite{schreiber2017pomegranate}, cross validating over the number of valence bins (2, 4, or 8) and the number of Gaussian components (1, 2, or 3). Valence predictions were computed by using the Viterbi algorithm to infer the most likely sequence of valence bins, followed by a simple moving average across every 5 time-steps.

Like SVR, the HMM does not perform well either, achieving a maximum performance of .04 $\pm$ .11 on the Validation Set and .04 $\pm$ .15 on the Test Set. Although the HMM is a time-series model, it is still unable to perform well given the complexity of the current dataset. This is likely due to the limited capacity of the model, which assumes that the input features are not correlated within each Gaussian component (i.e. diagonal covariance), and that each bin of valence ratings corresponds to only one underlying hidden state. We provide the SVR and HMM model results as baselines and to facilitate comparison with previous papers. We move next to discussing two state-of-the-art models.

\subsection{Using Long Short-Term Memory Networks}

\label{sec:Modeling:LSTM}

\begin{figure}[!bt]
\centering
\includegraphics[width=.8\columnwidth]{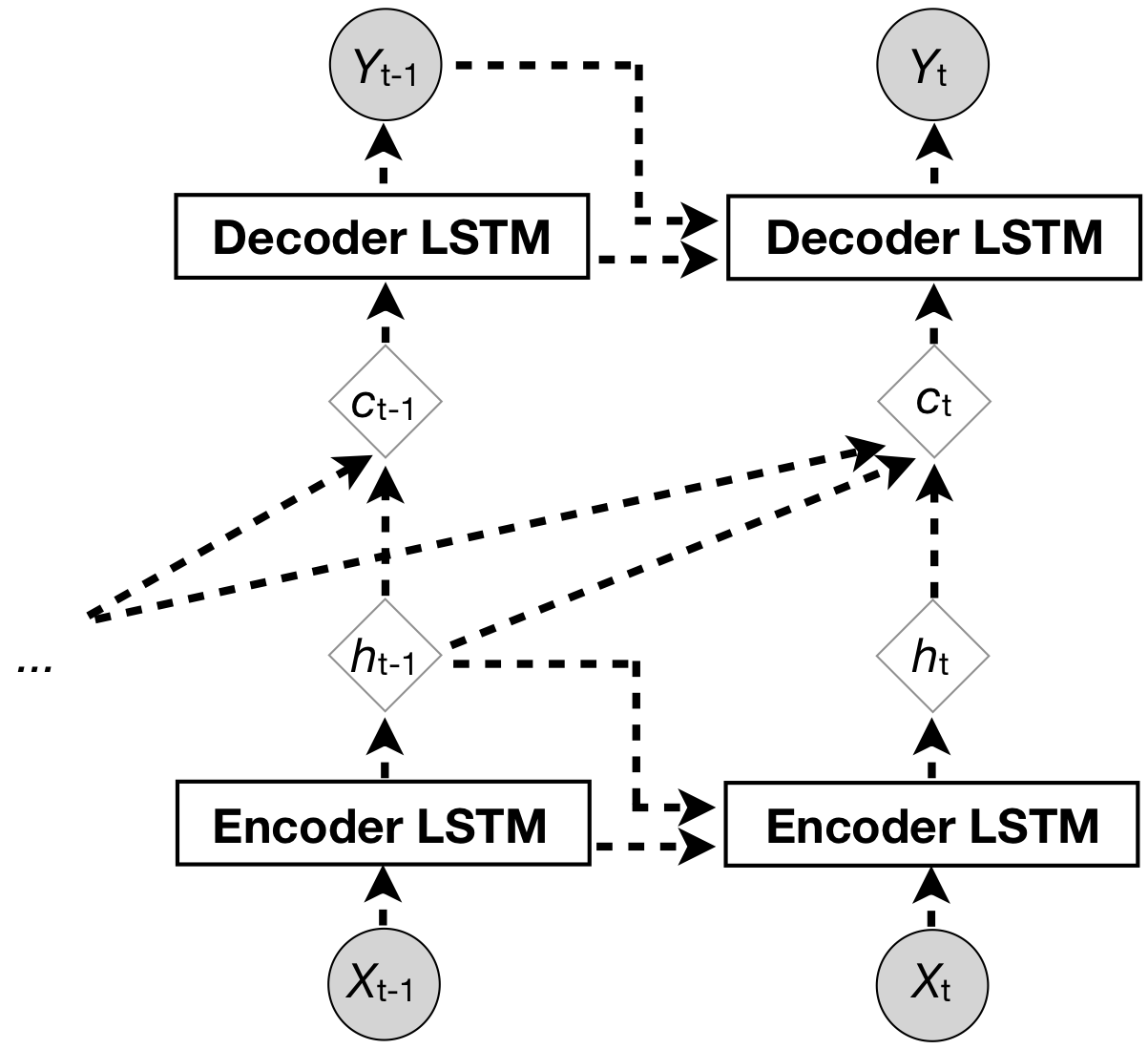}
\caption{Illustration of the Encoder-Decoder LSTM model. $X_t$ is a multimodal feature vector, and $Y_t$ is a real-valued valence rating. The first layer puts $X_t$ through an LSTM to encode a hidden layer representation $h_t$. The local attention layer of length $l$ computes a set of $l$ attention weights (Eqn. \ref{eqn:LSTM_attention}), and computes the context variable $c_t$ as a linear combination of the hidden units (Eqn. \ref{eqn:LSTM_context}). The context vector $c_t$ is then fed into a second, LSTM decoder layer to provide the final output $Y_t$. 
}
\label{fig:models-LSTM}
\end{figure}

As we noted, the Long Short-Term Memory (LSTM) deep neural network is one of the most popular and successful discriminative approaches in time-series emotion recognition. It provides a flexible framework that can learn general nonlinear functions from multimodal input features ($X_t$) to an emotion output ($Y_t$, in our case, valence).
Here, we implement an Encoder-Decoder LSTM, which consists of two LSTM layers, with a local attention layer in between (Fig. \ref{fig:models-LSTM}). This encoder-decoder architecture has previously been applied to predict sequences in other domains (e.g., machine translation \cite{sutskever2014sequence}). 


First, the ``encoder" LSTM layer takes in the input sequence $X_1,\ldots,X_t$ and computes hidden states $h_1,\ldots,h_t$. 
Next, we compute a local attention layer \cite{bahdanau2015neural, luong2015effective} using a single-hidden-layer neural network (or Multilayer Perceptron, MLP) with an attention window of length $l$. This means that, at time $t$, we compute a set of $l$ attention weights which are then used to weight the hidden states at the current and previous $l-1$ timesteps, to give a context vector $c_t$:
\begin{align}
    \text{Encoder Layer:} \qquad \qquad h_t &= \text{LSTM}(X_{1:t}) \\
    \{a_{t-l+1}, \ldots, a_t\} &= \text{MLP}(X_t) \label{eqn:LSTM_attention} \\
    c_t &= \sum_{j=0}^{l-1} a_{t-j} h_{t-j} \label{eqn:LSTM_context}
\end{align}

Finally, we added a second ``decoder" LSTM to predict the output $Y_{t}$, from the current context vector $c_t$ and the previous time-step $Y_{t-1}$. During training, we used ``teacher-forcing" \cite{williams1989learning} with a ratio of 50\%, which means that with 50\% probability on the training cases, the decoder
LSTM was fed the actual value at the previous time step $Y_{t-1}$, while on the remainder, the LSTM used its predictions on the previous time-step $\hat{Y}_{t-1}$.
\begin{align}
    \text{Decoder Layer:}\quad \hat{Y}_t &= \text{LSTM}(c_t, \hat{Y}_{t-1})
\end{align}



We used the mean squared error of the predictions (i.e., $\text{MSE}(\hat Y_{1:T}, Y_{1:T}) = \sum_{t=1}^{T} (\hat{Y_t} - Y_t)^2$) as the loss function to be minimized. We trained the LSTM with an initial dropout layer (on the input embeddings) of 0.1, which helps to regularize the learnt weights and help prevent overfitting \cite{srivastava2014dropout}, and with an attention window of $l=3$.


\subsubsection{LSTM Results}





Our LSTM performed the best using the Text features, achieving a CCC of .38 $\pm$ .29 on the Validation set and .40 $\pm$ .32 on the Test set (Table \ref{tab:SummaryOfResults}). Our LSTM model also does well with the Text and Visual features on the Test set, with a similar CCC of .40 $\pm$ .33, although it did not do well for this modality combination on the Validation set. As expected, our LSTM performs significantly better than the baseline SVR and HMM models on the Test set, across all modalities (linear mixed-effect models regressing CCC on model, with random intercepts by video and modality; LSTM$-$SVR, $t=9.78$, $p<.001$; LSTM$-$HMM, $t=10.1$, $p<.001$), 
and also comparing the best-performing best-performing LSTM (Text-only, and Text-Visual) with the best-performing SVR and HMM (paired $t$-tests; all $p$'s$<.001$).

When we compare our LSTM model results with the human benchmark, we find that, on the Test Set, the performance of the LSTM with Text-only features is not significantly different from the human benchmark (paired $t$-test: $t(38)=1.87$, $p=.07$). This is also true for the LSTM with Text and Visual features ($t(38)=1.79$, $p=.08$). 



One limitation of our current LSTM models is that we do not leverage the ability of neural networks to extract features directly from the raw data. For example, many previous models use a CNN on the raw images to extract visual features (e.g., \cite{kahou2015recurrent, brady2016multi}), rather than calculating visual features separately as we did here. The weights of such a CNN will be modified during training, which ``optimizes" the feature extraction process for this particular task. We chose not to do that here to have the same input features across all models to facilitate comparison, although we think that learning feature extraction from raw input will likely improve the performance of the LSTM models.

\subsection{Using Variational Recurrent Neural Networks}

\label{sec:Modeling:VRNN}

\begin{figure}[!bt]
\centering
\includegraphics[width=.6\columnwidth]{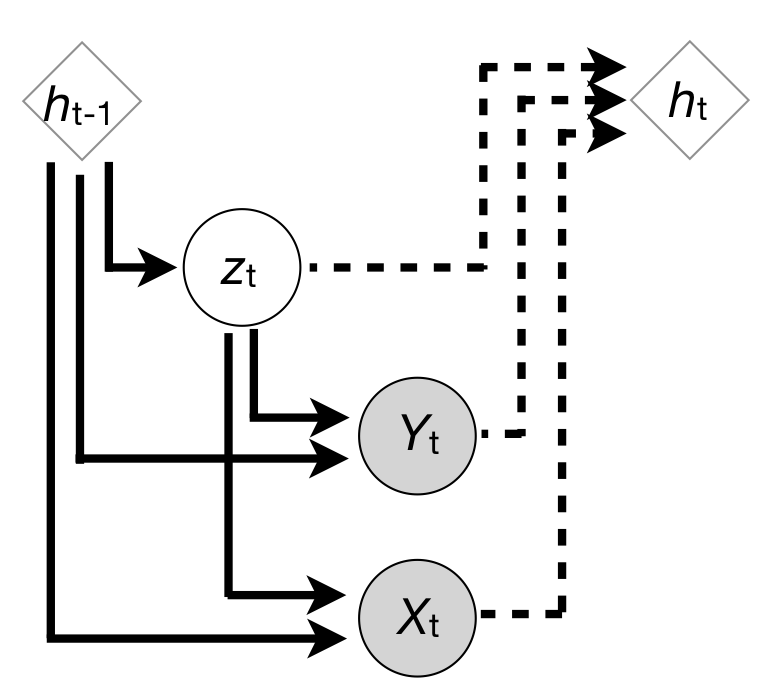}
\caption{Graphical structure of the multimodal Variational RNN (VRNN), adapted from \cite{chung2015recurrent, wu2018multimodal}. The hidden state at the preceding time step $h_{t-1}$ parameterizes all of the distributions at the current time step $t$. First, we estimate the posterior distribution $Q(z_t|X_t, Y_t ; h_{t-1})$ given the true inputs and outputs $X_t$ and $Y_t$, and sample $z_t$ from the posterior. Then we sample $\hat{X}_t$ and $\hat{Y}_t$ from the generating distribution $P(X_t, Y_t|z_t)$ to compute a reconstruction loss. Finally, we compute the recurrence to $h_t$ using the sampled $z_t$ and the observed $X_t$, $Y_t$ (replacing $X_t$ with $\hat{X}_t$ only if $X_t$ is missing, and similarly for $\hat{Y}_t$). We use a solid line to indicate ``causal flow" (as in a graphical model), and dashed lines to indicate a deterministic computation.
}
\label{fig:models-VRNN}
\end{figure}

The LSTM is excellent at learning mappings from the inputs $X_t$ to the outputs $Y_t$, but otherwise does not encode any other assumptions about the data. By contrast, adding latent variables to the model might allow us to account for implicit sources of variation (e.g., speaker-dependent attributes, differences in narrative style or theme, or inter-video differences), which might help us to generalize better across different videos. Thus, we wanted to see if combining a generative component into an RNN may result in better performance on the valence prediction task. 
%
One way to do this is to build a generative model of the inputs $X_t$ and the outputs $Y_t$, modeling them as generated from some lower-dimensional latent state $z_t$. By training the model to accurately predict both $X_t$ and $Y_t$, it could then automatically learn a good latent representation $z_t$ that captures the aforementioned sources of variation. If the model learns to map particular dimensions of $z_t$ onto these sources of variation, it could then go on to learn that some of them are irrelevant for predicting emotion, thereby allowing the model to generalize well across videos.

With this rationale, we implemented a multimodal Variational Recurrent Neural Network (VRNN; Fig. \ref{fig:models-VRNN}). We adapted the VRNN, proposed by \cite{chung2015recurrent}, to handle multiple modalities, by using a method from the (non-time-series) Multimodal Variational Autoencoder \cite{wu2018multimodal}. In our model, at each time step, we sample the latent variable $z_t$ from the approximate posterior $Q(z_t|X_t,Y_t)$, which is parameterized by the hidden state at the previous time step $h_{t-1}$, with parameters $\mu$, $\sigma$ learnt using deep networks. We follow \cite{wu2018multimodal} and assume a Gaussian prior $P(z_t)$ on the latent space, as well as Gaussian posteriors $Q(z_t|X_{t,m})$ for each input modality $X_{t,m}$ ($1\!\leq\!m\!\leq\!M$ where $M$ is the number of modalities); the full posterior $Q(z_t|X_t,Y_t)$ is then a product of Gaussians (itself a Gaussian).
\begin{align}
    z_t &\sim Q(z_t|X_t, Y_t) \nonumber \\
    &= P(z_t)\ Q(z_t|Y_t) \textstyle\prod_{m=1}^M Q(z_t|X_{t,m}) \\
    \text{where }
    & P(z_t) = \mathcal{N}\left( \mu_{z_t}, \sigma_{z_t} \right), \nonumber \\
    & Q(z_t|Y_t) = \mathcal{N}\left( \mu_{z_t|Y_t}, \sigma_{z_t|Y_t} \right),
    \nonumber \\
    & Q(z_t|X_{t,m}) = \mathcal{N}\left( \mu_{z_t|X_{t,m}}, \sigma_{z_t|X_{t,m}} \right), \nonumber 
    \\
    \text{and }
    &\mu_{z_t}, \sigma_{z_t} = \text{MLP}(h_{t-1}), \nonumber \\
    &\mu_{z_t|Y_t}, \sigma_{z_t|Y_t} = \text{MLP}(Y_t, h_{t-1}), \nonumber \\
    &\mu_{z_t|X_{t,m}}, \sigma_{z_t|X_{t,m}} = \text{MLP}(X_{t,m}, h_{t-1}) \nonumber
\end{align}
Next, we reconstruct the multimodal inputs $\hat{X}_t$ and outputs $\hat{Y}_t$ from the sampled $z_t$; these likelihood distributions are also parameterized by $h_{z-1}$. Finally, the recurrence occurs by computing the next hidden state $h_t$ via a deterministic computation from $z_t$, $X_t$ and $Y_t$, parameterized by a Multilayer Perceptron. In the event that there is a missing input modality $m$ at time $t$, we use the reconstruction $\hat{X}_{t,m}$ in place of the unobserved inputs $X_{t,m}$ to compute $h_t$. Similarly, we replace $Y_t$ with $\hat{Y}_t$ if the former is missing.
\begin{align}
    \hat{X}_t &\sim P(X_t|z_t) = \mathcal{N}\left( \mu_{X_t}, \sigma_{X_t} \right) \\
    \hat{Y}_t &\sim P(Y_t|z_t) = \mathcal{N}\left( \mu_{Y_t}, \sigma_{Y_t} \right) \\
    h_{t} &= \text{MLP}(z_t, X_t, Y_t) \\
    \text{where } 
    &\mu_{X_t}, \sigma_{X_t} = \text{MLP}(z_t, h_{t-1}) \nonumber \\
    &\mu_{Y_t}, \sigma_{Y_t} = \text{MLP}(z_t, h_{t-1}) \nonumber
\end{align}
To train the VRNN, we maximize the Evidence Lower Bound (ELBO) used in variational inference \cite{hoffman2013stochastic, blei2017variational}, summed across all timesteps $t$:
\begin{align}
    \textstyle\sum_{t=1}^T \biggl[  &\mathbb{E}_{Q(z_t|X_t, Y_t)} \left[\alpha \log P(Y_t|z_t)\right] \nonumber \\ 
    +\ &\mathbb{E}_{Q(z_t|X_t, Y_t)}\left[ \textstyle\sum_{m=1}^M \lambda_m \log P(X_{t,m}|z_t)\right] \\
    -\ &\beta\ \text{KL}[Q(z_t|X_t, Y_t)||P(z_t)] \biggr] \nonumber
\end{align}
Here, $\alpha$, $\beta$, and $\lambda_m$ are weights balancing the importance of each ELBO term, and $\text{KL}[Q||P]$ is the Kullback-Leiber divergence between distributions $Q$ and $P$. By maximizing the ELBO, the network simultaneously learns better generating distributions $P(Y_t|z_t)$ and $P(X_t|z_t)$, while performing regularization by ensuring that the approximate posterior $Q(z_t|X_t, Y_t)$ does not diverge too far from the prior $P(z_t)$.

During training, we gradually increase the weights $\alpha$ and $\beta$ from zero as we increase the number of epochs. This allows the network to first learn how to reconstruct the inputs $X_t$ by improving $P(X_t|z_t)$, before eventually placing more emphasis on both reconstructing the outputs $Y_t$ and regularizing the network. We also scale each $\lambda_m$ inversely with the dimensions of each input modality $m$, ensuring that reconstruction of that modality is not favored simply because it has more feature dimensions.

\begin{table}[!t]
    \scalebox{0.95}{
    \setlength\tabcolsep{2.5pt}
    \begin{tabular}{@{}lrrrrrrr@{}}
    \toprule
    \multirow{2}{*}{\textbf{Model}} & \multicolumn{7}{c}{\textbf{Modalities}} \\ \cmidrule(l){2-8} 
     & \multicolumn{1}{c}{A} & \multicolumn{1}{c}{T} & \multicolumn{1}{c}{V} & \multicolumn{1}{c}{AT} & \multicolumn{1}{c}{TV} & \multicolumn{1}{c}{AV} & \multicolumn{1}{c}{ATV} \\ \midrule
    \multicolumn{8}{c}{Validation CCC (Std. Dev.)} \\ \midrule
    SVR & .05 (.11) & .07 (.15) & .04 (.13) & .07 (.13) & .07 (.14) & .04 (.11) & .07 (.14) \\
    HMM & .03 (.09) & .04 (.11) & .04 (.15) & .03 (.12) & .04 (.12) & .02 (.06) & .01 (.11) \\
    LSTM & .10 (.30) & \textbf{.38 (.29)} & .12 (.30) & .08 (.29) & .28 (.30) & .10 (.24) & .14 (.29) \\
    VRNN & .11 (.26) & \textbf{.43 (.32)} & .11 (.24) & .32 (.31) & .24 (.30) & .14 (.30) & .17 (.26) \\
    Human & \multicolumn{1}{c}{--} & \multicolumn{1}{c}{--} & \multicolumn{1}{c}{--} & \multicolumn{1}{c}{--} & \multicolumn{1}{c}{--} & \multicolumn{1}{c}{--} & .47 (.15) \\ \midrule
    \multicolumn{8}{c}{Test CCC (Std. Dev.)} \\ \midrule
    SVR & -.02 (.13) & .08 (.16) & -.01 (.13) & .07 (.15) & .06 (.14) & -.01 (.11) & .06 (.14) \\
    HMM & .02 (.07) & .02 (.14) & .01 (.18) & .00 (.12) & .04 (.15) & .01 (.08) & .01 (.11) \\
    LSTM & .14 (.28) & \textbf{.40 (.32)} & .17 (.32) & .09 (.32) & \textbf{.40 (.33)} & .16 (.28) & .15 (.23) \\
    VRNN & .15 (.23) & \textbf{.42 (.32)} & .14 (.25) & .35 (.29) & .30 (.32) & .17 (.36) & .24 (.37) \\
    Human & \multicolumn{1}{c}{--} & \multicolumn{1}{c}{--} & \multicolumn{1}{c}{--} & \multicolumn{1}{c}{--} & \multicolumn{1}{c}{--} & \multicolumn{1}{c}{--} & .50 (.12) \\ \bottomrule
    \end{tabular}
    }
    \vspace{3pt}
    \caption{Summary of model results. Modalities---A: Audio, T: Text, V: Visual. Human: mean CCC between an individual human rater and the EWE of all other human ratings (described in Section \ref{sec:Modeling:HumanBenchmark}). 
    For the LSTM and VRNN, we indicate the best performing modality combinations in bold.
    }
    \label{tab:SummaryOfResults}
\end{table}

\begin{figure*}[!bt]
\centering
\includegraphics[width=.9\textwidth]{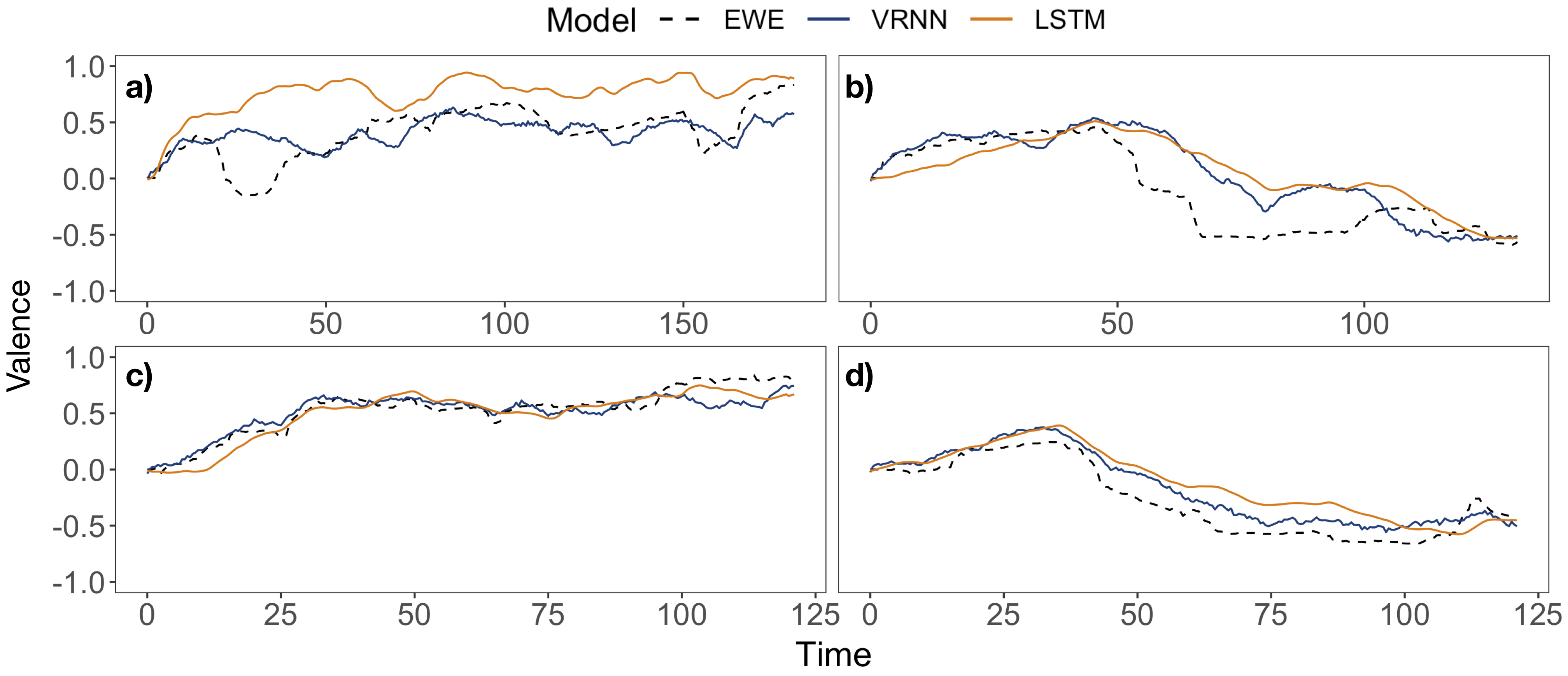}
\caption{Sample predictions of the best-performing model-modality combinations (LSTM: Text + Visual features; VRNN: Text features) compared with the EWE ratings (dashed black line). All plots shown are on videos from the Test set. (a) is the same video as Fig. \ref{fig:video_examples}, top.}
\label{fig:results}
\end{figure*}

\subsubsection{VRNN Results}

Overall, the VRNN well, performing the best with only Text features, achieving a CCC of .43 $\pm$ .32 on the Validation set and .42 $\pm$ .32 on the Test set. 
The performance of the VRNN is not statistically different with the performance of the LSTM, whether it is across all modalities (using a linear mixed-effect models regressing CCC on model, with random intercepts by video and modality, $t=1.50$, $p=.13$) or comparing only the best-performing modalities (LSTM-TV vs. VRNN-T; paired $t$-test, $p=.68$).
The performance of the best-performing VRNN is also not significantly different with the human benchmark ($t(38)=1.68$, $p=.10$).

Compared to the LSTM models, the VRNN theoretically models different sources of variability using the latent variable $z_t$. We predicted from our own qualitative impressions of the SENDv1 dataset that being able to account for different sources of variability would be critical to performance. However, although the VRNN does well, it did not do significantly better than the LSTM. 

\section{Discussion} \label{sec:Discussion}

In order to build artificial intelligence that understands human emotions, researchers must overcome the challenge of modeling emotion dynamics. In this paper, we address one piece of that puzzle---time-series emotion recognition---and offer a comprehensive review of contemporary time-series modeling approaches that are used or can be used productively in affective computing. We present a rich naturalistic dataset, the first version of the Stanford Emotional Narratives Dataset (SENDv1), designed precisely for multimodal, time-series emotion recognition. Finally, we report the results of several baseline and state-of-the-art models on the SEND, as a starting point for future work.

\subsection{The first version of the SEND}
A significant barrier to emotion-sensing AI is the lack of large, high-quality corpora for model training. To accomplish real-time emotion inference on real-world situations (``in the wild"), affective computing models need to first be trained on dynamic, multimodal, and naturalistic stimuli, but which are also well-controlled, i.e., captured in context, with a high signal-to-noise ratio. Our new corpus, the SENDv1, attempts to provide such a dataset. The paradigm was designed to create a minimally-constraining context, limiting undesirable noise, while still allowing for the naturalistic unfolding of emotion expression over time. 

Despite being a modestly-sized corpus, with N=193 video clips in the current version, the SENDv1 holds several advantages over readily available video stimuli such as excerpts from movies or YouTube videos. First, film clips or any acted media are staged, and therefore, not naturalistic. Such media do not capture genuine personal experience but instead, the actors' \emph{expectations} of experience, which are often exaggerated \cite{scherer2003vocal, gunes2008lab}. Though millions of ``in-the-wild" clips made by amateurs can be found on livestreams, video-logs, or websites like YouTube, this great quantity comes with a significant trade-off in quality. Videos may have poor lighting and audio, or exhibit large variations in framing and pose; and it is not always possible to know if the emotional expressions were staged or exaggerated. 

Furthermore, corpora collected or scraped from the Internet often lack a ``ground truth"; That is, the person expressing emotion in the clip did not provide self-reports as to what they were feeling. Even if these videos are annotated by online volunteers, such reports would not capture the ``ground truth" with regard to the personal experience of the person in the clip. Although we report results trained on the EWE calculated from independent observers, our dataset also contains self-reported ratings by the target in the video, as well as physiological measurements, trait, and demographic information, which may prove useful in building individualized models.

The SENDv1 data set has a high signal-to-noise ratio that is desirable for training machine learning models. We minimize undesirable variance in background noise and lighting by having only one person speaking in front of a black background with no distractions, while increasing desirable variance such as: (a) the \textit{diversity of targets} in the stimuli with respect to gender identity, race and ethnicity, 
communicative style, and age; and (b) the \textit{diversity of content}, with respect to the topics, places, people, and events discussed in the videos. Limiting videos to a single storyteller allowed us to study naturalistic expression with minimal noise. This approach is ideal for ``personal assistant" AI or AI for therapy applications, and also serves as a benchmark from which to build models that can understand dialogue between two or more people. 
It is still important for affective computers to  tackle truly ``in-the-wild" scenarios: this dataset focuses on distilling and focusing on some complexities (diversity of content and targets) over others (different contexts, arbitrary lighting and views).

We intend to extend the SEND in the following ways: First, we intend to augment the dataset with more videos, via a growing, international team of collaborators, increasing the diversity in age, ethnic, and other demographic variables, and even collecting content in more diverse languages \cite{jospeREVISIONcontribution}. 
Second, we intend to collect more varied moment-by-moment ratings (e.g., of discrete emotion ratings, or appraisal ratings), as the current rating scale simply measures emotional valence---the most-important principal component of emotions, but it is by no means exhaustive. 
We hope that this first version of the SEND is but the first of a cumulative set of resources for affective computing and psychology researchers, and that the SENDv1 and future extensions will enable more sophisticated, human-like AI.

\subsection{Modeling}
In its current form, the SENDv1 has proven to be a useful data set for training emotion recognition models; however, there is still room for future work on the modeling, in order to best extract and integrate the rich information from multiple modalities. For example, all our best-performing models used only one or two modalities, and future research could examine how to better integrate multimodal information to improve performance: We found in a recent investigation \cite{wu2019attending} that state-of-the-art models with simple concatenation fusion do poorly on multimodal inputs on the SENDv1, and required more sophisticated fusion methods to better integrate multiple modalities.

In particular, we find that, on our dataset, models trained on the linguistic features perform the best. This should not be surprising, because \emph{a priori}, we intuitively expected that the most important predictor of the emotions in a narrative would be the linguistic content of the narrative. 
However, we did not use very sophisticated linguistic features---we used a Bag of Words with GloVe features for each time window, which results in an averaged word vector. Importantly, these features may capture some semantic meaning in each window, but likely do not capture any narrative elements, such as the arc, climax, and resolution of a story. Representing and understanding narratives remains a challenging state-of-the-art problem in Natural Language Processing. From an affective computing perspective,  linguistic features should ideally capture how people subjectively \emph{interpret} events, which is an important precursor to emotions.

To reiterate this point in a broader context, the manner in which the majority of affective computing conceptualizes emotion understanding is primarily via emotion recognition from observable cues. That is, an affective computer ``understands" what a user is feeling if the affective computer perceives and processes behavioural cues like the user's facial expressions, and produces an output of what the user is feeling. This is a difficult task, due to the large complexity of how emotions are expressed in face, voice, and other modalities, and as we mentioned, the field has made much progress on this front \cite{zeng2009survey, poria2017review}. This assumption is also encapsulated in the discriminative time-series approaches we reviewed, which is to find the best (statistical) mapping from the behavioral cue data to an emotion label or rating. 

However, from a psychological perspective, emotion recognition is just one of the many ways that people can understand someone's emotions \cite{ong2015affective, ong2019computational}. People understand how others' emotions arise as \emph{responses to events} in the world---including via subjectively evaluating the significance of the event, as in Appraisal Theories of emotion \cite{ellsworth2003appraisal, ortony1988cognitive, ong2019computational}---or how emotions dynamically vary in interpersonal interactions \cite{mesquita2014emotions}. More broadly, a theoretically-driven approach would suggest building a causal model of how emotions arise, how they vary over time, and how they result in behavior \cite{ong2019applying}, and use these causal models as a basis for emotion understanding. This is the assumption behind the generative approach (and event-based approaches \cite{du2016recurrent, linderman2014discovering}, which we did not cover here), which posits a causal data-generating process. 
These perspectives offer exciting potential for capturing and modeling affective dynamics.

There is still much work to be done: We note that the generative models we presented still do not capture events and appraisals, and still rely on behavioral cues. Furthermore, as mentioned, our use of word-vector representations for linguistic cues does not identify real-world events (e.g. ``I had a breakup") or the subjective appraisals that subsequently accompany these events.
We think a fruitful set of future directions include integrating existing models of what constitutes an emotionally-relevant event (e.g., from computational appraisal theories and first-person emotion architectures \cite{ortony1988cognitive, marsella2009ema}) into machine-learning models, perhaps via a generative or event-based approach.

More generally, a causal model-based approach is also applicable beyond multimodal time-series emotion recognition to \emph{longitudinal} emotion understanding---that is, understanding emotions over the course of many sessions. For example, a medical robot that sees a patient once every few months would need to maintain a longitudinal record of what were the events that happened to the patient---diagnosis and continual treatment records, progression of the medical condition---in order to decide how best to affectively respond to the patient. Empathic doctors naturally do this, especially if the medical condition is sensitive (e.g., terminal or incurable), and even if there are long gaps between patient visits. Such longitudinal emotion understanding can be thought of as a generalized version of the time-series problems we discussed in this paper: The observations (patient-robot interactions) may be irregularly spaced, and may be driven by other ``events" such as test results and other medical information. We hope that some of the ideas from the time-series models we discussed will also prove useful in longitudinal modeling.

In conclusion, time-series emotion recognition is a crucial component of affective computing. In this paper, we have outlined several challenges of---as well as several state-of-the-art solutions to---capturing dynamics in emotion recognition. We hope that this discussion will inspire more ambitious, theoretically-driven modeling using diverse combinations of approaches.

\ifCLASSOPTIONcompsoc
  \section*{Acknowledgments}
\else
  \section*{Acknowledgment}
\fi

The authors would like to thank Emma Master, Kira Alqueza, Michael Smith, and Erika Weisz for assistance with the project, and Noah Goodman for discussions about modeling. This work was supported in part by the A*STAR Human-Centric Artificial Intelligence Programme (SERC SSF Project No. A1718g0048), a Stanford IRiSS Computational Social Science Fellowship to DCO, and NIH Grant 1R01MH112560-01 to JZ.


\ifCLASSOPTIONcaptionsoff
  \newpage
\fi


\bibliographystyle{IEEEtran}
\bibliography{Biblio-TAC}


\begin{IEEEbiography}[{\includegraphics[width=1in,height=1.25in,clip,keepaspectratio]{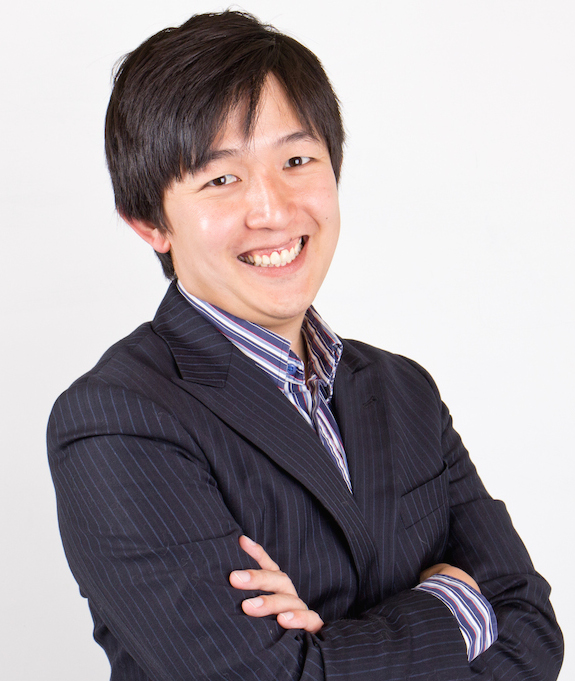}}]{Desmond C. Ong} is an Assistant Professor of Information Systems and Analytics at the National University of Singapore. He holds a concurrent appointment as a Research Scientist with the A*STAR Artificial Intelligence Initiative. Desmond received his Ph.D. in Psychology and M.Sc. in Computer Science from Stanford University, and he graduated with a B.A. in Economics (\textit{summa cum laude}) and Physics (\textit{magna cum laude}), with minors in Cognitive Studies and Information Science from Cornell University. His research interests include building computational models of emotion and mental state understanding, using a mix of human behavioral experiments and modeling approaches like probabilistic modeling and machine learning. He is a member of the IEEE Computer Society.
\end{IEEEbiography}

\begin{IEEEbiography}[{\includegraphics[width=1in,height=1.25in,clip,keepaspectratio]{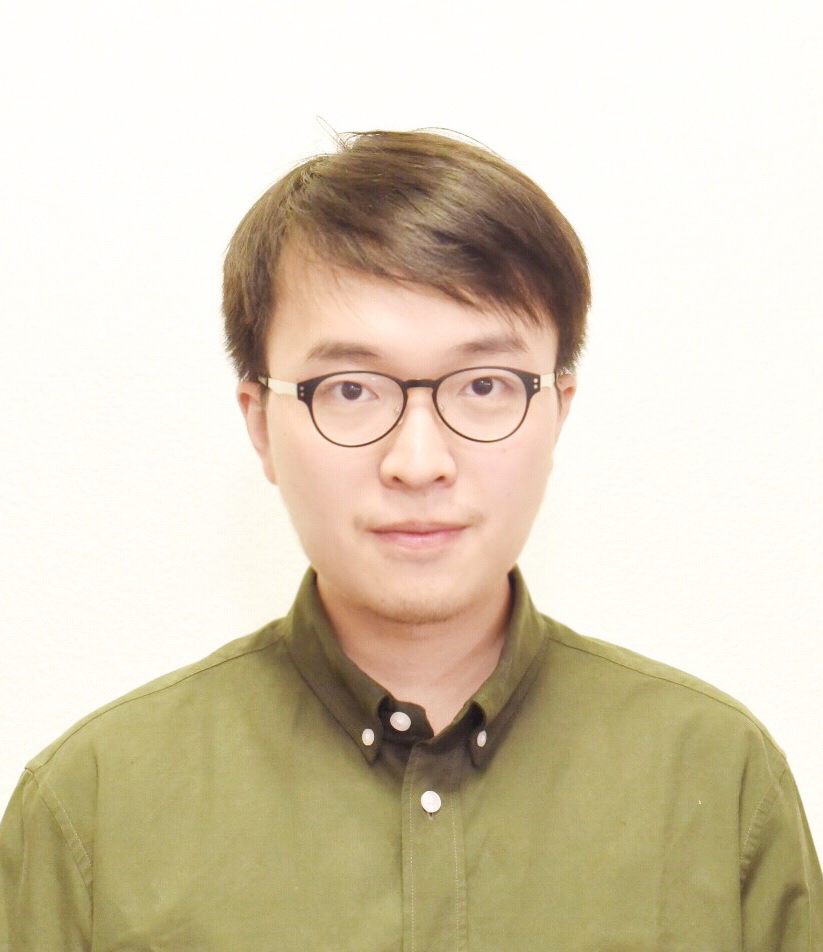}}]{Zhengxuan Wu}
is completing his M.Sc. in Management Science and Engineering with a concentration in Computational Social Science at Stanford University. He graduated with a B.S. in Aerospace Engineering (\textit{magna cum laude}) and Mechanical Engineering (\textit{cum laude}) from Case Western Reserve University. He also received a M.Sc. in Computer Science from the University of Pennsylvania. His research interests include studying the interplay between emotion and cognition with the applications of machine learning algorithms and computational modeling.
\end{IEEEbiography}

\begin{IEEEbiography}[{\includegraphics[width=1in,height=1.25in,clip,keepaspectratio]{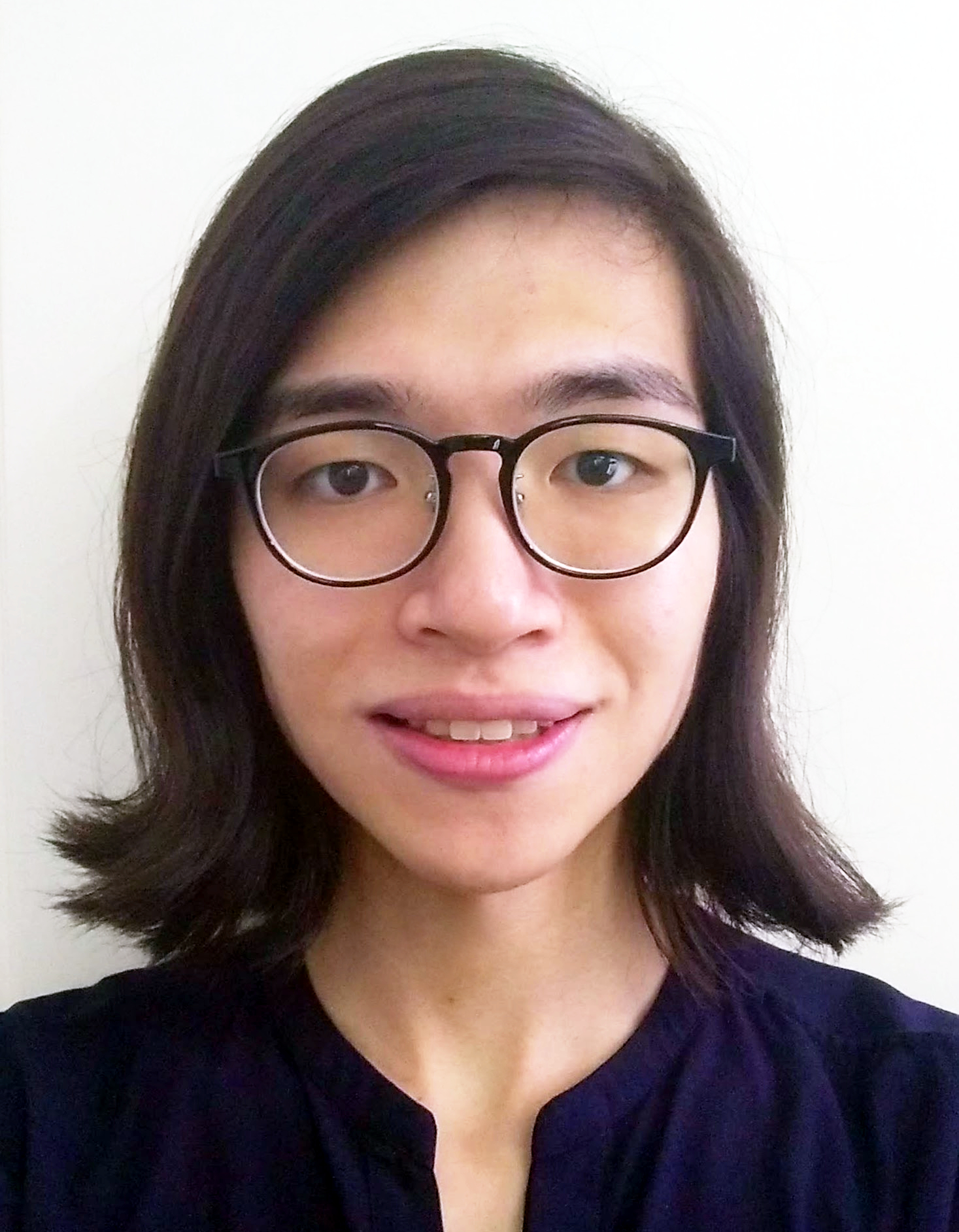}}]{Tan Zhi-Xuan} received a B.S. in Electrical Engineering and Computer Science (\textit{magna cum laude}) from Yale University in 2018, and is currently a Ph.D. student at the Massachusetts Institute of Technology. Prior to starting graduate school, Xuan was a Research Engineer with the A*STAR Artificial Intelligence Initiative. Their research interests include computational modeling of human moral psychology, as well as using cognitively-inspired approaches to build AI systems that can better understand and conform to people's intentions, goals, norms, and values. \end{IEEEbiography}

\begin{IEEEbiography}[{\includegraphics[width=1in,height=1.25in,clip,keepaspectratio]{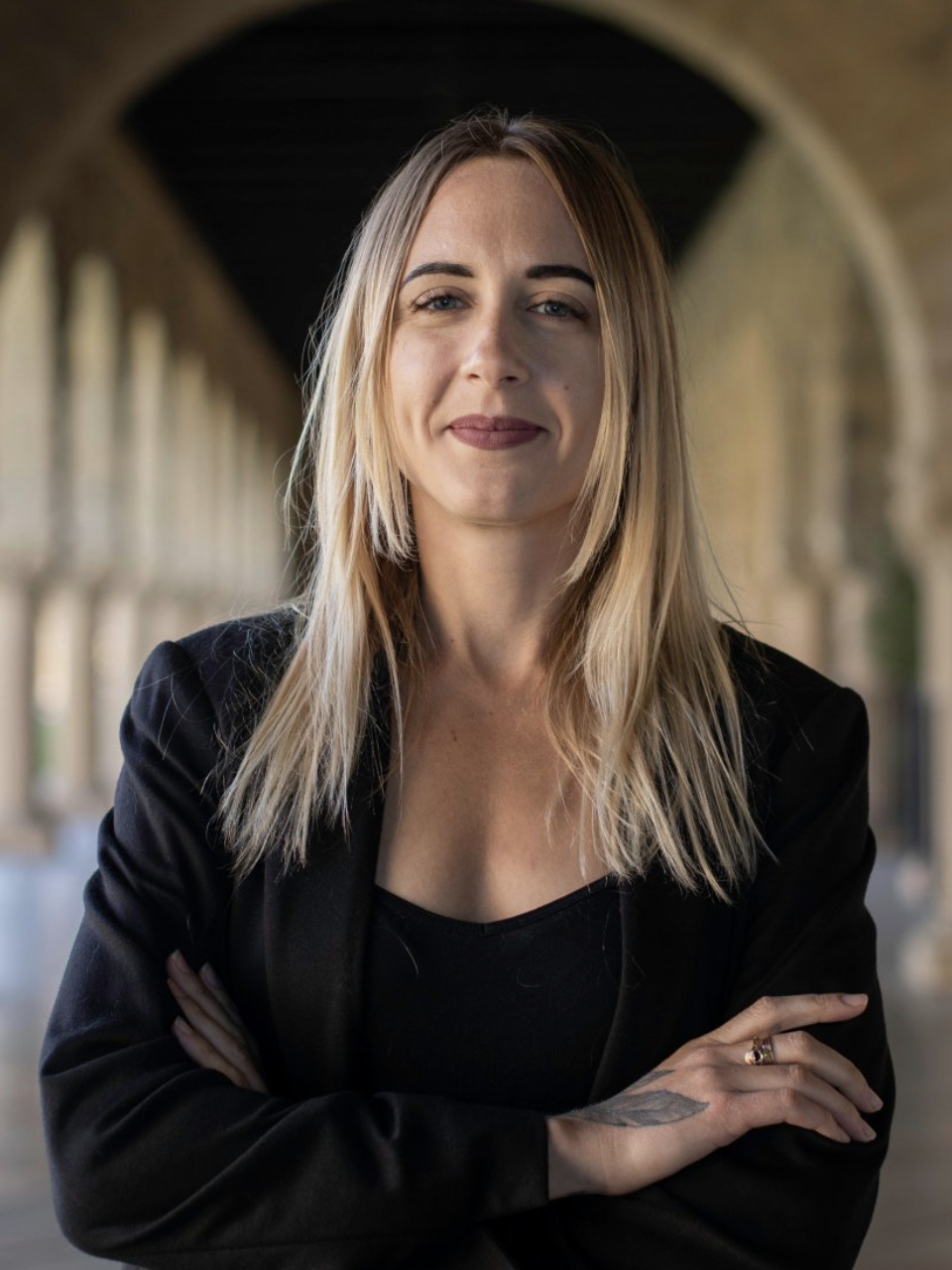}}]{Marianne Reddan} obtained her Ph.D. from the laboratory of Tor Wager at the University of Colorado Boulder in a combined degree program that intersects Cognitive Science and Psychology and Neuroscience. She is currently a postdoctoral researcher in the Department of Psychology at Stanford University. Her research interests include modeling the neural and physiological processes underlying emotion expression and modification. She uses machine learning to develop signatures of emotion expression which can then be targeted through behavioral interventions to improve quality of life.
\end{IEEEbiography}

\begin{IEEEbiography}[{\includegraphics[width=1in,height=1.25in,clip,keepaspectratio]{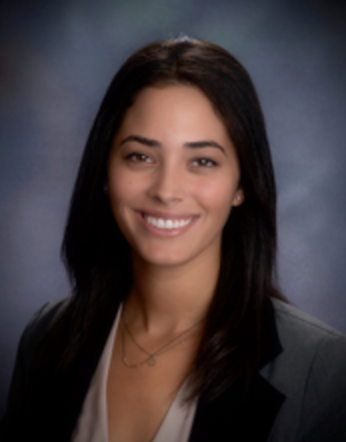}}]{Isabella Kahhale} received her B.S. in Cognitive \& Brain Sciences (\textit{summa cum laude}) from Tufts University in 2017, and then worked as a research coordinator with Professor Jamil Zaki in the Stanford Social Neuroscience Lab. She is currently pursuing a PhD in a joint Clinical-Developmental Psychology program at the University of Pittsburgh under the mentorship of Professor Jamie Hanson. Her research interests include empathy gaps with respect to underserved communities and the impact of early life stress on antisocial behaviors.
\end{IEEEbiography}

\begin{IEEEbiography}[{\includegraphics[width=1in,height=1.25in,clip,keepaspectratio]{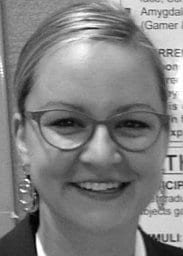}}]{Alison Mattek}
obtained her Ph.D. in Psychological and Brain Sciences from Dartmouth College in 2017, and worked as a postdoctoral researcher in Psychology at Stanford University. She is currently a postdoctoral researcher in the Department of Psychology at the University of Oregon. Her research interests include emotion and motivation.
\end{IEEEbiography}

\begin{IEEEbiography}[{\includegraphics[width=1in,height=1.25in,clip,keepaspectratio]{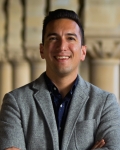}}]{Jamil Zaki} is an Associate Professor of Psychology at Stanford University, where he has been on the faculty since 2012. He received his Ph.D. in Psychology from Columbia University in 2010 and did his postdoctoral work at Harvard University. 
He has won numerous awards, such as the 2017 Sage Young Scholar Award, a 2016 Early Career Award from the Society for Social Neuroscience, a 2015 NSF CAREER Award, and a 2015 Janet T. Spence Award for Transformative Early Career Contribution and a 2013 Rising Star award, both from the Association for Psychological Science. His research interests include empathy and emotion understanding.
\end{IEEEbiography}

\end{document}